\documentclass[runningheads]{llncs}

\usepackage{eccv}

\usepackage{eccvabbrv}
\usepackage{lmodern}
\usepackage{graphicx}
\usepackage{booktabs}

\usepackage[pagebackref,breaklinks,colorlinks,citecolor=eccvblue]{hyperref}

\usepackage[utf8]{inputenc} %
\usepackage[T1]{fontenc}    %
\usepackage{url}            %
\usepackage{booktabs}       %
\usepackage{amsfonts}       %
\usepackage{amsmath}
\usepackage{nicefrac}       %
\usepackage{microtype}      %
\usepackage{xcolor}         %
\usepackage{graphicx}
\usepackage{enumitem}
\usepackage{nicematrix}
\usepackage{multirow}
\usepackage{gensymb}        %
\usepackage{array}          %
\usepackage{tabularx}       %
\newcolumntype{C}{>{\centering\arraybackslash}X} %
\usepackage{array}
\usepackage{etoolbox}

\newtoggle{inlinecomments}
\toggletrue{inlinecomments}  %
\iftoggle{inlinecomments}{
}{
}

\begin{document}

\title{Making Images from Images:\\ Interleaving Denoising and Transformation}

\author{%
  Shumeet Baluja\inst{1}
  \and
  David Marwood\inst{1}
  \and
  Ashwin Baluja\inst{2}
}

\authorrunning{S.~Baluja et al.}

\institute{Google DeepMind \and Northwestern University}

\maketitle
\begin{abstract}

Simply by rearranging the regions of an image, we can create a new
image of any subject matter.  The definition of regions is user
definable, ranging from regularly and irregularly-shaped blocks, 
concentric rings, or even individual pixels.  Our method extends and
improves recent work in the generation of optical illusions by
simultaneously learning not only the content of the images, but also
the parameterized transformations required to transform the desired images
into each other.  By learning the image transforms, we allow any
source image to be pre-specified; any existing image (\emph{e.g.} the
\emph{Mona Lisa}) can be transformed to a novel subject.  We formulate
this process as a constrained optimization problem and address it
through interleaving the steps of image diffusion with an energy
minimization step.  
Unlike previous methods, increasing the number of regions actually makes the problem easier and improves results.
We demonstrate our approach in both pixel and latent spaces.  Creative
extensions, such as using infinite copies of the source image and
employing multiple source images, are also given.

\end{abstract}

\section{Introduction and Background}

\begin{figure}
  \centering
  A.
 \includegraphics[height=1in]{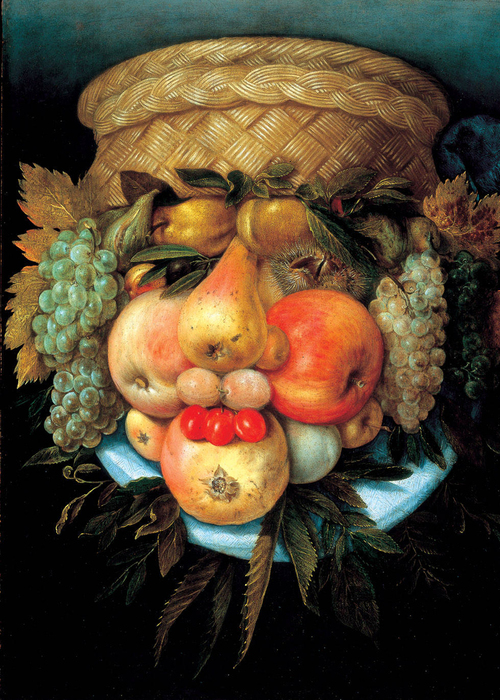}
 ~~
 ~~
 ~~
 B.
 \includegraphics[width=1in]{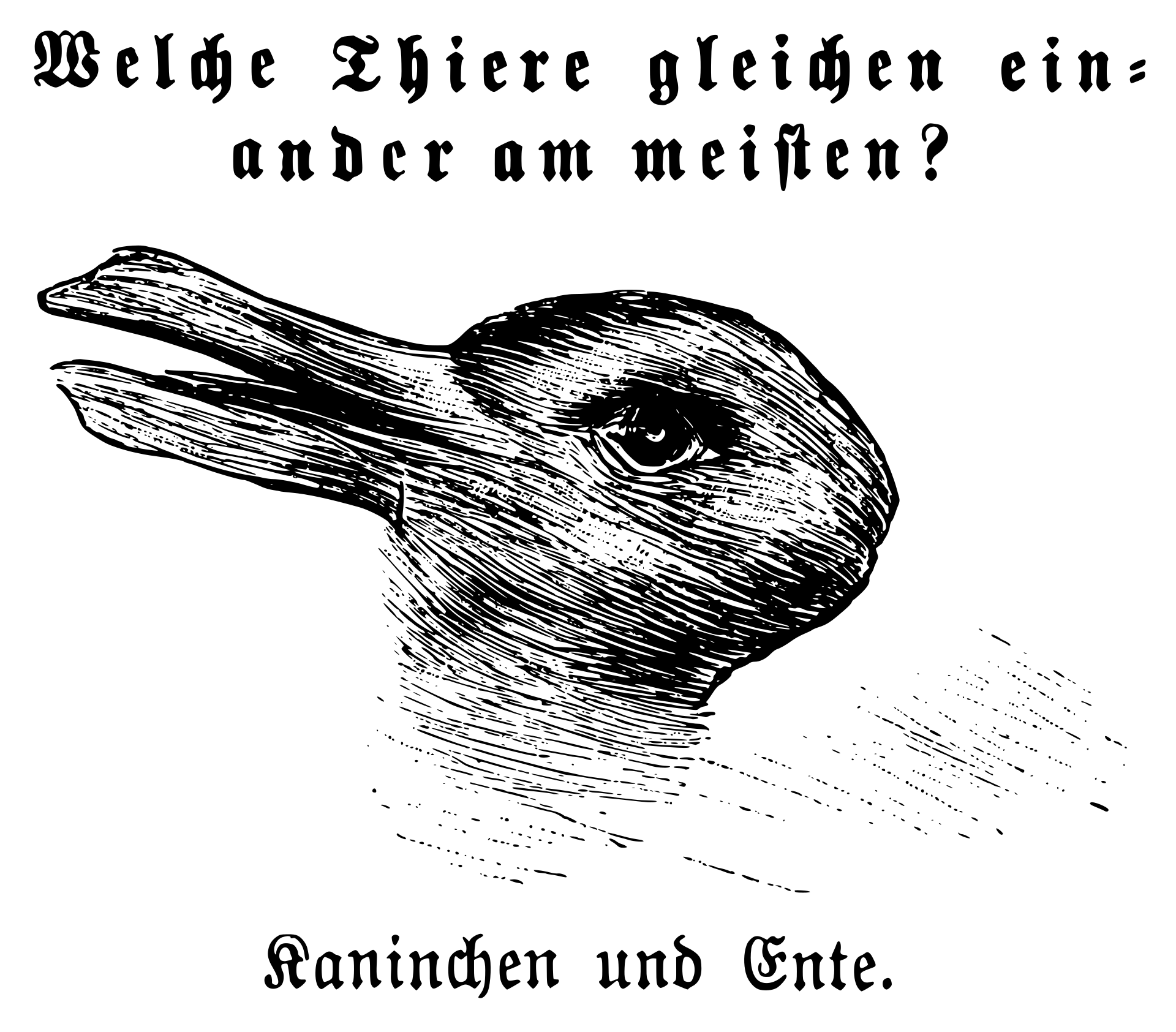}

 \caption{Classic examples of optical illusions.  (A)  G.Arcimboldo's \emph{Fruit Basket} (1590) that shows a face when upright, and a fruit basket when upside-down.  (B)  Depending on the orientation, this image appears either as a duck or a rabbit. %
 }
 \label{historic}
\end{figure}

The generation of images that change their appearance under specific
transforms, such as rotations, tile rearrangement, and flips, has
captivated the minds of visual artists for centuries --- exemplified
by Salvador Dalí and M. C. Escher. 
Often grouped broadly as optical illusions, two prominent early examples  include
the work of Giuseppe Arcimboldo (1590), shown in
Figure~\ref{historic}A, and the classic example of the rabbit and duck
illusion, shown in Figure~\ref{historic}B.  Beyond their artistic appeal as
intriguing visual puzzles, these and other optical illusions have been
studied for understanding human perception, including intrinsic biases
and fundamental cognitive
processes~\cite{lo2011investigation,das2018role,de2015visual}. ~\cite{ngo2023is,jaini2024intriguing}
found that modern large-scale image/text embedding systems are also
susceptible to illusions (a type of adversarial
example~\cite{goodfellow2014explaining,foolHumans,Hendrycks_2021_CVPR}).

Within the computer vision community, the quest to create complex
visual illusions that can trick the human visual system has
encompassed many approaches.  For example, \emph{hybrid images}
combine the high frequencies from one image with the low frequencies
from another to create a hybrid image that appears as one object when
viewed close-up and another when viewed from
afar~\cite{hybridImages}. Other work includes creating solids that can
be interpreted as different objects depending on the viewing
angle~\cite{mitra2009shadow,hsiao2018multi} and generating the
illusion of motion without movement~\cite{freeman1991motion}.

\begin{figure*}[t]
\centering
\newcommand{\sfs}{0.61in}
\newcommand{\sfss}{0.61in}
\newcommand{\sfsss}{0.61in}
\newcommand{\leftCol}{0.5in}

\newcommand{\ssfs}{1.35in}
 \begin{tabular}{p{0.5in}ccc}

\begin{minipage}[t]{\leftCol}
\centering
\scriptsize
  \vspace{-0.35in}
 Source\\
 Image
\end{minipage} &

\begin{minipage}[t]{\ssfs}
\centering
\tiny
  \includegraphics[height=\sfs]{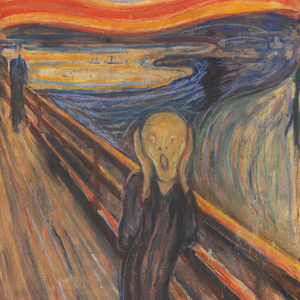}  \\
  \emph{The Scream}\\ by
  Edvard Munch
\end{minipage} &

\begin{minipage}[t]{\ssfs}
\centering
\tiny
  \includegraphics[height=\sfs]{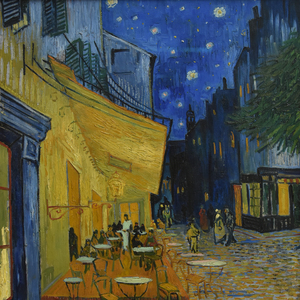}\\
  \emph{Cafe~Terrace~at~Night}\\ 
  by Van Gogh
\end{minipage} &

\begin{minipage}[t]{\ssfs}
\centering
\tiny
  \includegraphics[height=\sfs]{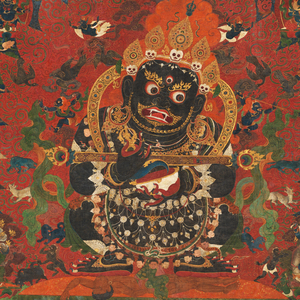} \\
  \emph{Mahalaka}\\ by unknown(1400AD)
\end{minipage} \\

  \midrule
  
 \begin{minipage}[t]{\leftCol}
 \centering
  \vspace{-0.35in}
  \scriptsize $64\times64$\\
  tiles

 \end{minipage} &

 \begin{minipage}[t]{\ssfs}
  \centering

\includegraphics[height=\sfs]{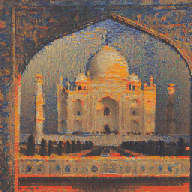}~\includegraphics[height=\sfs]{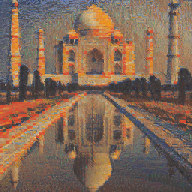}\\
 \tiny Taj Mahal
 \end{minipage}
&
 \begin{minipage}[t]{\ssfs}
  \centering 
 \includegraphics[height=\sfs]{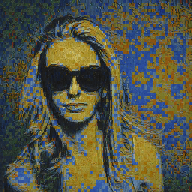}
  \includegraphics[height=\sfs]{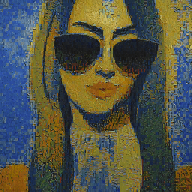}\\
  \tiny Woman with Sunglasses
 \end{minipage}  
&
 \begin{minipage}[t]{\ssfs}
  \centering 
 \includegraphics[height=\sfs]{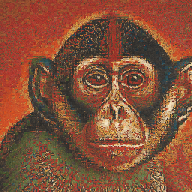}
  \includegraphics[height=\sfs]{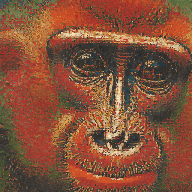}\\

  \tiny Monkey 
 \end{minipage} 
 \\
 
 \midrule
 \begin{minipage}[t]{\leftCol}
 \centering
   \vspace{-0.35in}
 \scriptsize $32\times32$\\
 tiles
 \end{minipage} &
 \begin{minipage}[t]{\ssfs} 
  \centering 
 \includegraphics[height=\sfss]{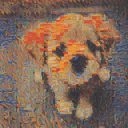}
 \includegraphics[height=\sfss]{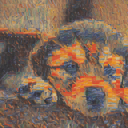}\\
  \tiny Cute Puppy
 \end{minipage}  
&
 \begin{minipage}[t]{\ssfs}
  \centering 
 \includegraphics[height=\sfss]{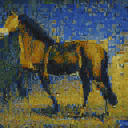}
 \includegraphics[height=\sfss]{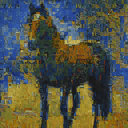} \\
  \tiny Horse
 \end{minipage}  
 &
 \begin{minipage}[t]{\ssfs} 
  \centering 
  \includegraphics[height=\sfss]{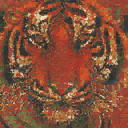}
  \includegraphics[height=\sfss]{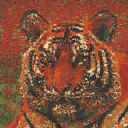} \\ 
  \tiny Tiger
 \end{minipage}   
 \\
 
  \midrule

\begin{minipage}[m]{\leftCol}
\centering
  \vspace{-0.35in}
 \scriptsize $16\times16$ \\
 tiles
 \end{minipage} &
 \begin{minipage}[t]{\ssfs}  
  \centering 
 \includegraphics[height=\sfsss]{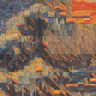}
 \includegraphics[height=\sfsss]{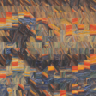}\\
  \tiny Great Wave %
 \end{minipage}    
&
 \begin{minipage}[t]{\ssfs} 
  \centering 
 \includegraphics[height=\sfsss]{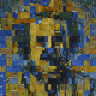}
 \includegraphics[height=\sfsss]{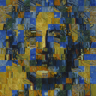}\\
  \tiny Einstein
 \end{minipage}    
 &
 \begin{minipage}[t]{\ssfs}  
  \centering 
  \includegraphics[height=\sfsss]{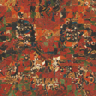}
  \includegraphics[height=\sfsss]{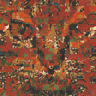} \\ 

  \tiny Cat
 \end{minipage}     

\end{tabular}

 \caption{Through simple tile permutations, a source image can be converted to a new image of any subject matter.  Both the permutation and the content are learned simultaneously; the images created are suited to the tiles available for the composition.  Examples with three famous paintings are shown.  Each is converted into 3 different subjects (two results are shown for each).  The number of tiles that the source is divided into is  $64\times64$, $32\times32$ and $16\times16$ (top to bottom).  With our approach, as the number of tiles grows, the \emph{easier} it is our for system to produce compelling results; the opposite is true for state-of-the-art alternate systems.}
 \label{fig:teaser}

\end{figure*}

Recently, a wide variety of novel work has been conducted in
algorithmically creating \emph{photo-realistic} optical illusions.
The foundation of almost all of this work is the text-to-image
generative
models~\cite{mj,dhariwal2021diffusion,saharia2022photorealistic,song2020denoising,rombach2021highresolution,ramesh2021zero,croitoru2023diffusion,nichol2021glide,imagen2022}
that have become the de facto standard method to synthesize
high-quality images from text prompts.  To gain more fine-grained
control of the generation process, as is necessary for creating
illusions, a variety of methods have been proposed that either replace
or modify specific objects in the generated images or manipulate the
style of
generation~\cite{tumanyan2023plug,avrahami2022blended,nichol2021glide,Everaert_2023_ICCV}.
Samples of our work, shown in Figure~\ref{fig:teaser}, provides a
novel mechanism for controlling the generation of images.

In work close to ours, Tancik used diffusion to simultaneously denoise
multiple views of the same image~\cite{illusionDiffusion}; two
different user-specified subject were visible by rotating the
image. Geng \emph{et al.}\cite{geng2023visualanagrams} significantly
broadened those results by showing that any pre-specified
rearrangement of pixels is a suitable transformation that can be
captured by the simultaneous denoising processes.  They used multiple
prompts within a pixel-based diffusion system,
DeepFloyd~\cite{deepfloyd}, to create a single image that, through
pre-specified transforms, represented the multiple prompts.  An
alternative approach~\cite{burgert2023diffusion} used Stable
Diffusion~\cite{rombach2021highresolution} to create similar rotation
illusions as well novel overlay illusions in which multiple realistic
images are overlaid to reveal an image of a new subject matter.  In
2024, a broadened set of transforms was presented
by~\cite{geng2024factorized}.  This included automatically generated
photo-realistic hybrid images (of the type that were first presented
in 2006~\cite{hybridImages}), in which a single image appears as
different objects at different scales.  %

A limitation of all of the previous work is that it has been confined
to a narrow problem scope: multiple images were generated
simultaneously to ensure closely compatible content under the
pre-specified
transforms~\cite{liu2023compositional,geng2023visualanagrams,geng2024factorized,chen2024images,burgert2023diffusion}.
Let us consider a more difficult problem.  Imagine that we are given a
source image, such as the artwork in Figure~\ref{fig:teaser}(top row), and our
goal is to create images that depict entirely different subjects using
only tiles from the source. Using a static source image under the
constraints of a pre-specified transform (as in previous work) does not work.  
The result is already fully determined and is not dependent on the
prompt.

To tackle this problem, we must first create a method to allow the
transformation to be dynamically discovered.  Second, even with
dynamic transforms, we note that the constraints on the images to be
created are much tighter with a static source image than one that can
be co-created.  In this new formulation, we can only create the new
image with the tiles from the first; arbitrary pixel colors are not
possible.  Put another way, there is more freedom in finding
compatible pixels for both images when the pixel values are not
pre-determined. 

The closest previous work that bridges the studies with optical
illusion creation and our tightly constrained diffusion process is by
Bar-Tal et \emph{al.}~\cite{bartal2023multidiffusion}. There, the
constraints on multiple simultaneous diffusion processes are
maintained to seamlessly extend and blend multiple images.  This was
used for creating panoramic images and region-based content
modification.  Additionally, we note that~\cite{geng2024factorized}
also used real images as one target stage for their hybrid-image
illusions.  They exploited the visibility differences of high and low
frequency features at different resolutions - a very different
approach and goal than ours.

As will be described next, our method does not require a pre-specified
arrangement of tiles, but instead determines the optimal permutations
dynamically as the new image content is generated. This dynamic patch
matching technique employs the \emph{Hungarian
Method}~\cite{hungarian}, also known as the
\emph{Kuhn-Munkres}~\cite{munkres1957algorithms} algorithm, for
optimal assignment. The image transformation process is modeled as an
optimization problem with shifting constraints and is addressed by
alternating steps of image diffusion and energy minimization. We also
extend our method to cases employing infinite copies of the source
image, providing a powerful tool for generating diverse and visually
appealing images from a single starting image.

Finally, we will show how the general approach of interleaving diffusion
and energy minimization steps can be extended to other image
transformations beyond tile rearrangement.

\section{Diffusion and Constrained Optimization}
\label{dco}

To address the task of using a static source image for creating a new image, we first reformulate the more general task of simultaneously denoising two non-static images.  We employ a diffusion process~\cite{Ho2020denoising} common to text-to-image systems that denoises an image over $T$ steps, starting from pure noise: $x_T \sim \mathcal{N}(0,1)$. $t$ proceeds from $T$ down to $0$. 
$x_t$ is the input to step $t$ which produces $x_{t-1}$, an image with less noise, as the input to the next step.
$x_0$ is the final output of diffusion:  the fully denoised image.   
By incorporating deep neural networks  with classifier free guidance~\cite{classifierfreeguidance} to predict the noise at each step, the diffusion process gradually changes pure noise into an image described by a text prompt, $p$.

\begin{figure}[t]
  \centering
  \includegraphics[width=\linewidth]{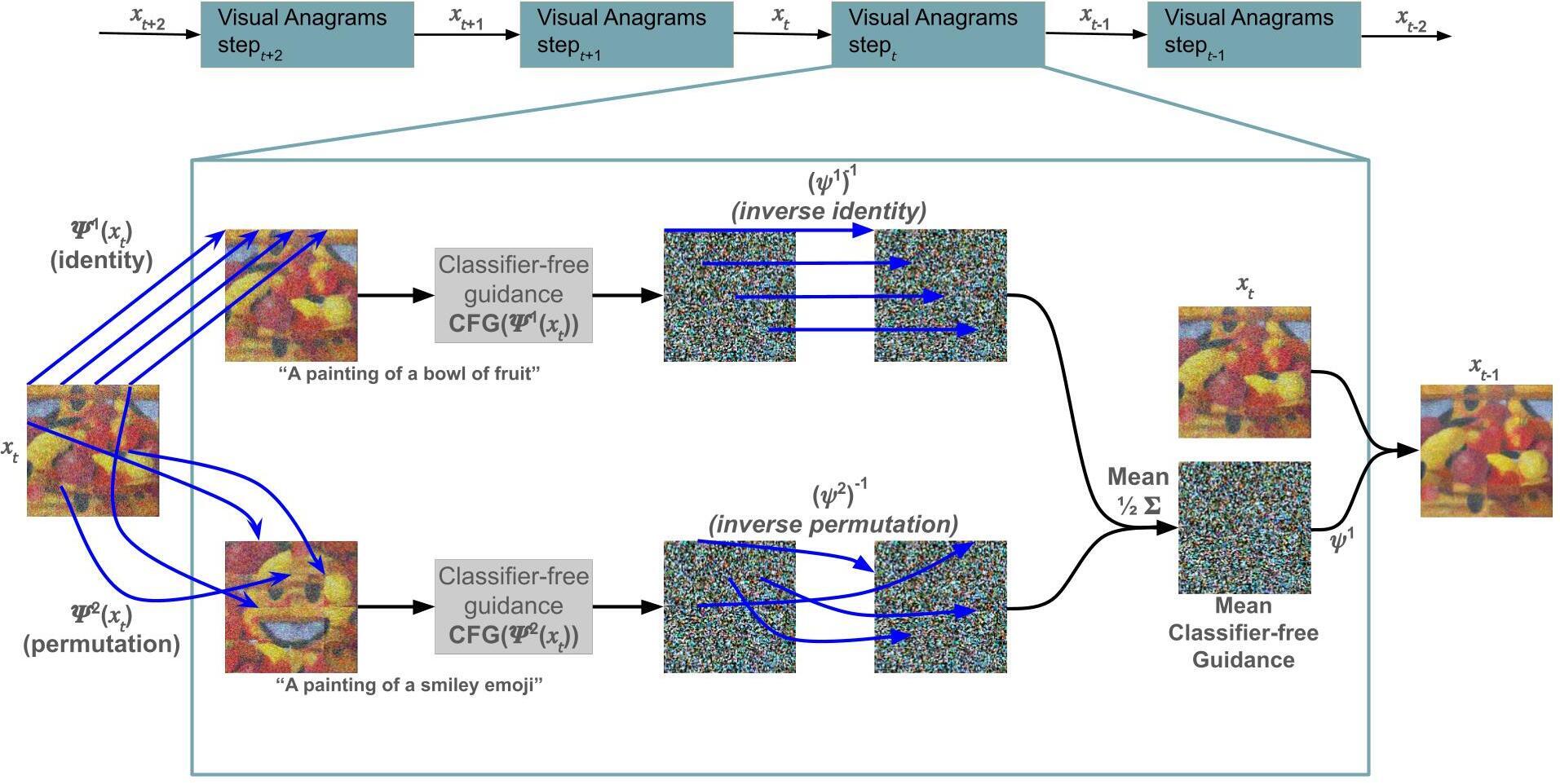}
 
  \caption{A description of a Visual Anagrams step, adapted from~\cite{geng2023visualanagrams}, using $N=2$. A single image that appears as a bowl of fruit can be subdivided into 4x4 square tiles and rearranged into a ``smiley face'' emoji.  To create this, two transforms of the same image (here $\psi^1$=identity and $\psi^2$ is a permutation of $4\times4$ tiles) are denoised simultaneously. Two diffusion processes use different prompts to create their per-pixel classifier-free guidance (CFG), represented here as an image. The CFGs are passed through their respective inverse transforms into a shared space and averaged before being applied to input $x_t$. Both images contribute to the guidance and remain synchronized.
  }
  \label{fig:visualAnagrams}
  \vspace{-0.1in}
\end{figure}

We base our system on Visual Anagrams~\cite{geng2023visualanagrams} which uses diffusion to synthesize $N$ images from $N$ different prompts such that the images at step $t$ are the same under the inverse of each transformation $\psi^i$, denoted $\psi^{i^{-1}}$.  The corresponding pixels in the images, as determined using $\psi^i$, are exactly the same. For simplicity, going forward, we will define the first of $N$ diffusion inputs to be the input image, $x_t$. The remaining diffusion inputs can be derived from the first as $\Psi^i(x_t) = \psi^i(\psi^{1^{-1}}(x_t))$ where $1 \le i \le N$ ($\Psi^1$ is identity)\footnote{This definition of $\Psi$ provides generality. If one is willing to restrict $\psi^1$ to identity then $\Psi^1(x_t) = \psi^1(x_t)$.}; see Figure~\ref{fig:visualAnagrams}. In~\cite{geng2023visualanagrams}, they \emph{a priori} select the transforms, $\psi$. To synchronize the $N$ parallel diffusion processes, the computed guidance for each prompt is averaged at each step.

For the majority of this study, we will concentrate on transforms, $\psi$, that divide the image into $M \times M$ tiles and permute the tiles intact rather than treating pixels individually.

\begin{figure}[t]
  \centering

 \tabcolsep=0.1cm
 \tiny
  \begin{tabular}{cc|cc|cc}
     \multicolumn{2}{c|}{\includegraphics[width=1.5in]{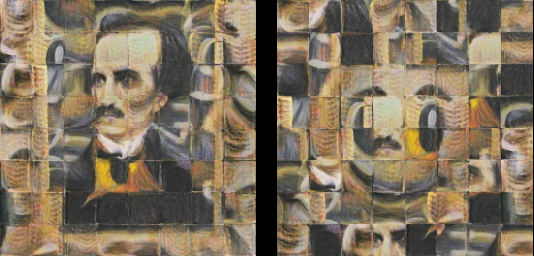}} &
     \multicolumn{2}{c|}{\includegraphics[width=1.5in]{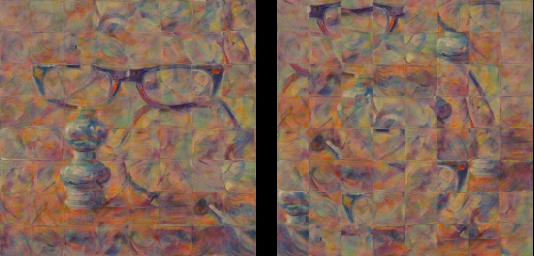}} &
     \multicolumn{2}{l}{\includegraphics[width=1.5in]{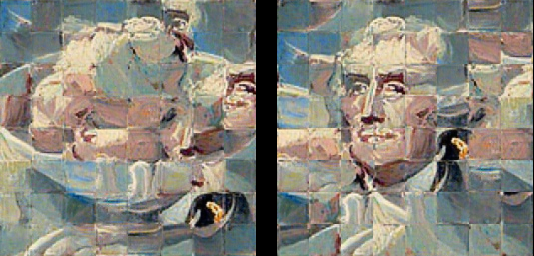}} \\
        
 Edgar Allan Poe&  Smiley emoji & 
 Glasses on a desk &  A spiral &
Bowl of ice cream & G. Washington \\
 \end{tabular}

  \caption{\footnotesize In each pair, each image is 192$\times$192 pixels. Each image can be subdivided into a grid of $8\times8$ (64 total blocks) of 24$\times$24 pixels.  For each pair of images, there is a permutation of the 64 blocks that transforms the left to right and vice-versa.  As in~\cite{geng2023visualanagrams}, the permutation is pre-specified and chosen randomly.  To show~\cite{geng2023visualanagrams} in the best light, these are the top performing, %
  as judged by their  CLIP scores (distance to prompt)~\cite{radford2021learning}.  All  prompts were preceded by ``A painting of''.}
     \label{fig:nonHungarianResults}
     \vspace{-0.1in}
\end{figure}

As a baseline, we show results obtained by using the approach in~\cite{geng2023visualanagrams} using $N=2$, see Figure~\ref{fig:nonHungarianResults}.  As in~\cite{geng2023visualanagrams}, $\psi^1$ is an identity function and $\psi^2$ is a randomly chosen tile permutation.  Also, to match their studies, we use the pre-trained diffusion model  DeepFloyd~\cite{deepfloyd}  that operates directly on pixels.    Our qualitative observations of the results match those in~\cite{geng2023visualanagrams}: prompts that allow more room for interpretation, such as ``a painting of..'' perform better than those with strict expectations, \emph{e.g.} ``a photograph of..''.  Additionally, the broader the query, the higher the likelihood of a pleasing result, \emph{e.g.} the pair (``a car'' and ``a tree'') will appear better than the pair (``a red Mustang'' and ``a willow tree'').

 \newcommand{\wff}{0.6in}

The most immediate limitation of~\cite{geng2023visualanagrams} is that the transformation function (the permutation) is static and selected prior to starting the process.  To elucidate the concern, consider the pair of prompts (``beach sunset'', ``basketball'').  It is easy to imagine that the color and shape of a basketball can be used for orange skies and a setting sun.  However, if the randomly chosen tile assignment is incompatible with such a layout, either a less likely image is generated or the objects in the image may appear incorrect.  Next, we address this problem.

\subsection{Simultaneously Learning Permutations and Images}
\label{sec:permutations}

It is impossible to know \emph{a priori} what the suitable permutations for a set of prompts will be.  We will develop a procedure that allows the transformation permutation to change such that the permutations are selected that best match the images.  Before proceeding, it is important to state the intrinsic ``chicken-and-egg'' problem. If the two final images were known, we could calculate the best permutation in advance.  However, the two images are being synthesized simultaneously along with their permutations. Therefore, we need to repeatedly re-optimize the permutation as the image is synthesized.

We find the permutation using \emph{dynamic matching}.  Consider two images, each composed as a grid of $M \times M$ non-overlapping tiles of equal size. The distance between tiles is their $L_2$ pixel distance, though more elaborate mechanisms such as perceptual similarity metrics can also be employed~\cite{amir2021understanding}.  The goal is to assign each tile in the first image to exactly one distinct tile in the second such that the total summed distances of assignments is minimized. This can be characterized as a complete bipartite graph $(S, T; E)$ where vertex sets $S$ and $T$ represent tiles in the first and second images and $E$ has an edge distance equal to the distance between its endpoint tiles. The solution is a perfect matching (a matching where each vertex has exactly one edge) that minimizes the total edge distance. This formulation will be used again in Section~\ref{sec:infiniteImages}.

To compute this, the tiles of each image are flattened to an array of length $M^2$ and the distances between every tile in the first image and every tile in the other image are arranged in a matrix, $D$. We permute the rows of $D$ to minimize its trace.
\begin{equation}
\min_P ~tr(PD)
\label{eq:dynamicMatching}
\end{equation}
where $P$ is a permutation matrix, a (0,1)-matrix with exactly one 1 in each row and at most one 1 in each column, with the same shape as $D$.  This assignment problem is optimally solved in polynomial time using the Kuhn-Munkres algorithm~\cite{hungarian,munkres1957algorithms}. 
$\psi(Q)$ implements the tile assignment $P$ on the pixels of $Q$.

We incorporate dynamic matching as part of the ``mainline'', the sequence of ``Visual Anagrams steps'' described so far.  Its role is to update the matching tiles between Visual Anagrams steps as the images are created.  Specifically, dynamic matching updates $\psi$ at step $t$, referred to as $\psi_t$ (and similarly $\Psi_t$), to $\psi_{t-1}$.
A naive approach is to simply recalculate the assignment in $\psi$ at each Visual Anagrams step; however, this was unstable -- likely because matching noisy images in the beginning (closer to $t=T$) resulted in unhelpful assignment changes. 

\begin{figure}[t]
 \centering
 \includegraphics[width=\linewidth]{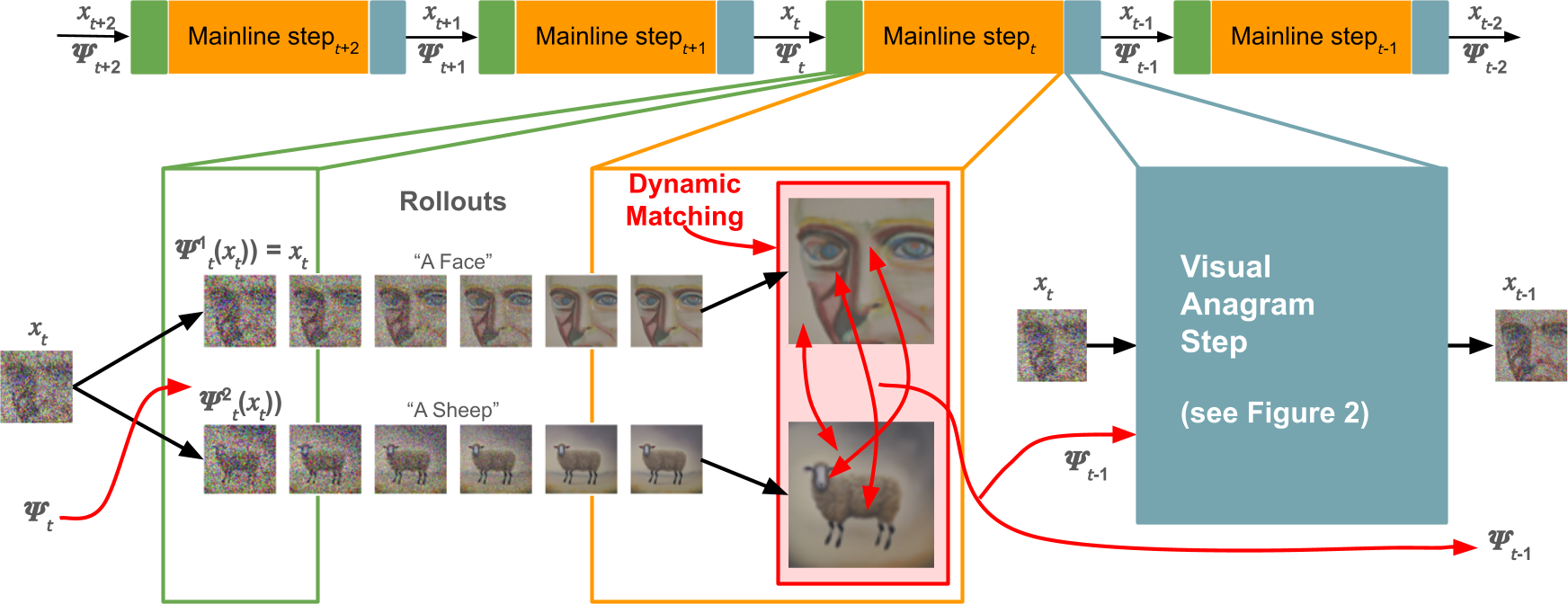}
 \caption{\small Incorporating dynamic matching between each diffusion step.  In the green rectangle, the input image is permuted and the rollout occurs.  In the orange rectangle, the images created at the end of the rollout are used to compute the new permutations through dynamic matching.}
 \label{fig:modifiedHungarianAnagram}

\end{figure}

To address this, 
we introduce a ``rollout'' at the beginning of each mainline step. A rollout is an \emph{entirely separate} diffusion process starting from $\Psi^i_t(x_t)$ and stepping forward $l$ steps as a form of lookahead approximation to the \emph{a posteriori} image for each prompt. (Numerous values were tested for $l$ and, experimentally, we found a lookahead of 5 steps was sufficient.) 
This approximation outputs an ``idealized'' image to show where each prompt \emph{would} lead if used in diffusion \emph{independently}. 
We then use dynamic matching \emph{on the idealized images} to compute new tile assignments; this produces the new $\psi^i_{t-1}$ and $\Psi^i_{t-1}(x_{t-1})$. 
As shown in Figure~\ref{fig:modifiedHungarianAnagram}, the $N$ rollouts take $\Psi^i_t(x_t)$ as input and the mainline step outputs $x_{t-1}$.

After the dynamic matching step, the idealized images are discarded;  only the new assignments are maintained. The Visual Anagram step proceeds as usual, using $\Psi_{t-1}(x_t)$ as input\footnote{More precisely, the diffusion process inputs are $\psi^i_{t-1}(\psi_t^{1^{-1}}(x_t))$.  However, any set of transforms $\psi$ can be equivalently restated without changing  $\psi^1$; thus, $\psi^1_t = \psi^1_{t-1}$. By substitution,  the diffusion process inputs are $\psi^i_{t-1}({\psi_{t-1}^1}^{-1}(x_t))$ which is just $\Psi^i_{t-1}(x_t)$.} and outputs $x_{t-1}$; see Figure~\ref{fig:modifiedHungarianAnagram}.

\begin{figure}[h]
  \centering
 \includegraphics[width=3in]{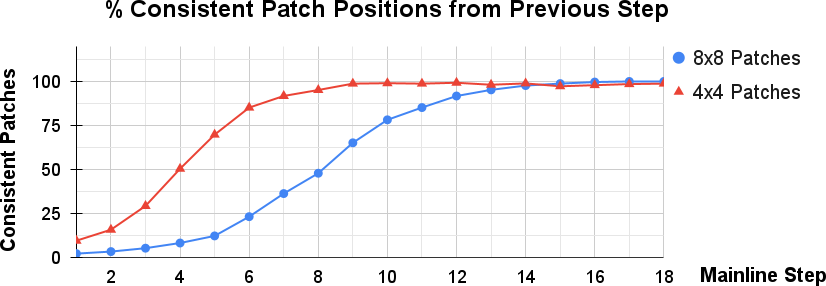}

  \caption{\small As the diffusion process continues, changes between successive permutations is reduced.  As the images become more consistent (as $t\rightarrow 0$) each $\psi_{t-1}$ looks more like $\psi_t$.    By iteration 15, the $8\times8$ tiles have ceased movement.  Movement ends by iteration 9 for the $4\times4$ tiles.  Average of 10 runs shown. }
   \label{fig:permutationMoves}

\end{figure}

It is interesting to examine, on a running system, how the dynamics of the tile permutations change across mainline steps.
Figure~\ref{fig:permutationMoves} shows that 
the number of changes in tile assignments rapidly reduces.
This occurs because, as the mainline steps proceed, the output images of each of the rollouts get closer to the final output (and thereby change less between iterations); therefore,  the optimal matches between the rolled-out images cease shifting.

\begin{figure}[t]
  \centering
  
 \tabcolsep=0.1cm
 \tiny
  \begin{tabular}{cc|cc|cc}
 
       \multicolumn{2}{c|}{\includegraphics[width=1.5in]{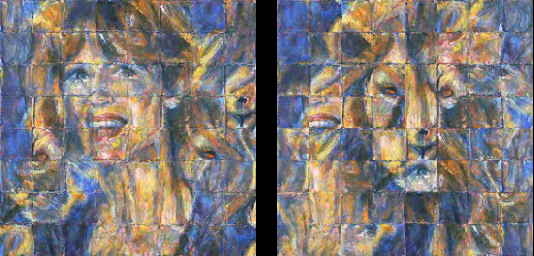}} &         
   \multicolumn{2}{c|}{\includegraphics[width=1.5in]{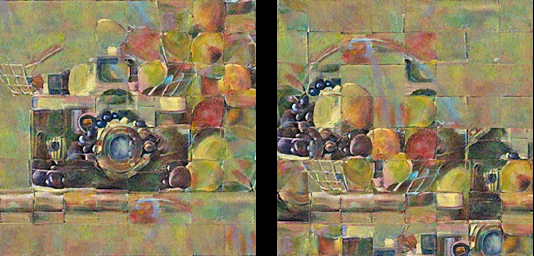}} &
     \multicolumn{2}{c}{\includegraphics[width=1.5in]{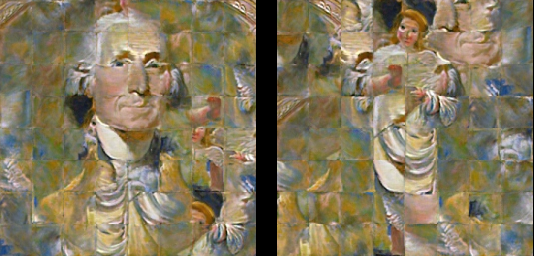}} \\
   Woman & Lion  & Old Camera & Bowl of Fruit & G. Washington&  Angel in White\\
& \\

 \multicolumn{2}{c|}{\includegraphics[width=1.5in]{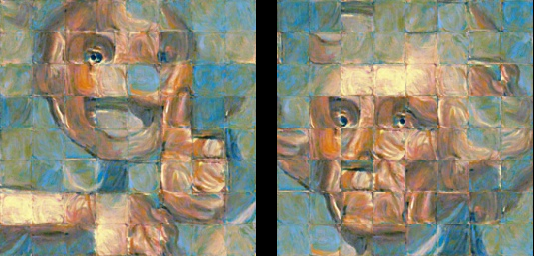}} &
     \multicolumn{2}{c|}{\includegraphics[width=1.5in]{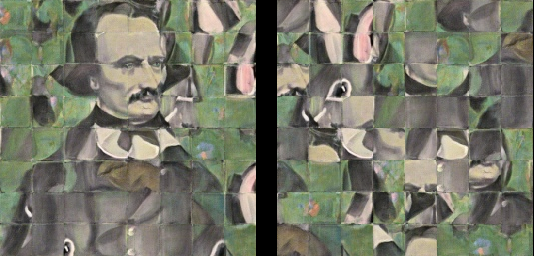}} &      
     \multicolumn{2}{c}{\includegraphics[width=1.5in]{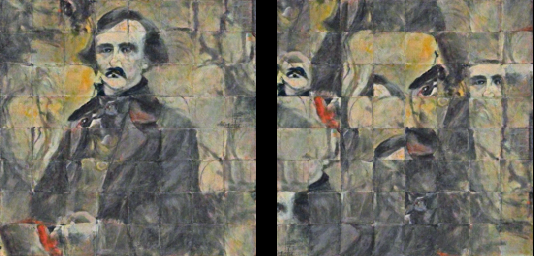}} \\    
      Thumbs-up emoji & Thomas Jefferson & Edgar Allan Poe & A Bunny &  Edgar Allan Poe & Bearded Man \\
   
 \end{tabular}

  \caption{\small The transformation is dynamically determined as the image is created. Note that the same prompt (Edgar Allan Poe), when paired with different prompts, yields different images because of the constraints imposed by the pairing.}
  
      \label{fig:hungarianResults}
\end{figure}

Qualitative results with dynamic matching are shown in Figure~\ref{fig:hungarianResults}. 
To provide a quantitative measure of quality,  1200 trials were conducted, half with dynamic matching and half without.  Each trial synthesized an image based on two prompts (randomly chosen nouns).  The distance of the resulting images to their respective prompts was measured in the CLIP embedding space~\cite{radford2021learning}.
For $4 \times 4$ permutations, dynamic matching averaged 22.39 distance and 22.43 without. The statistical difference, measured with the non-parametric Wilcoxon paired significance test, is ($p < 0.05$).  For $8 \times 8$ permutations, the results were 22.36 and 22.40, with significance ($p < 0.01$). 

In summary, by employing this novel matching between Visual Anagrams steps, we discover the tiles that are similar in appearance in the synthesized idealized images, and tie them together.  We no longer need to adhere to randomly chosen ties. Revisiting the ``basketball'' and ``beach sunset'' example and using dynamic matching, tiles that are orange in the separate images may be tied together, regardless of where the original random permutation placed them.  Next we show how this capability  allows us to tackle a new range of tasks.

\section{Using a Known Source Image}
\label{sec:fixed}

In the previous section, dynamic matching allowed us to generate multiple images from multiple prompts while progressively discovering the optimal tile assignments between them. Here, we exploit this new freedom to \emph {a priori} specify the tiles that all the images must contain.  The tiles can be taken from any ``source'' image;  if the \emph{Mona Lisa} is used, then the generated images will only use its exact tiles.

Starting with a source image changes the task from finding the RGB-pixel values  to rearranging the tiles of the source image to create a new image described by the specified prompt.  On first glance, this may seem like a trivial problem: given the \emph{Mona Lisa}, find the tiles that best match another image, such as an image of a dog.  However, this \emph{is not} what is being done here.  Instead, we are \underline{creating} an image of the dog that is well suited to the images allowed by the (\emph{e.g.} $16^2!$ for the $16\times16$ tile case) permutations of the \emph{Mona Lisa}'s tiles. 

\begin{figure}[t]
  \centering
  \includegraphics[width=\linewidth]{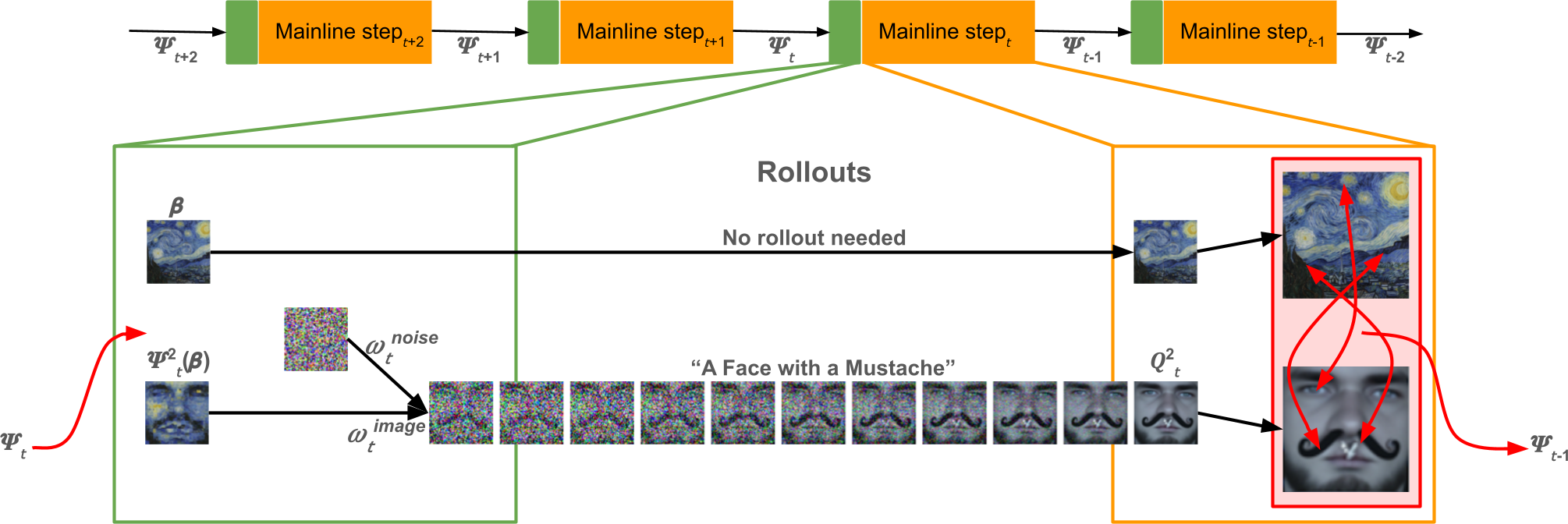}

  \caption{With a fixed image, $\beta$, the rollout procedure with an additional noising step is used to replace both the mainline and rollout procedures described in the previous section. In this example, an image of a man with a mustache is created by moving tiles around from Van Gogh's \emph{Starry Night}.
  Also, note that we no longer need to keep $x_t$; instead, $x_t$ is directly computable from $\Psi_t(\beta)$. }
  \label{fig:fixedDFAlgorithm}
\end{figure}

\newcommand{\eyew}{0.55in}
\begin{figure}[t]
  \centering
  \tabcolsep=0.08cm
\tiny
\begin{tabular}{c|cccc|cc|c}

  &
  
    \includegraphics[height=\eyew]{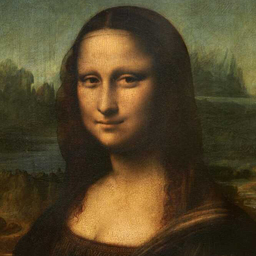}&
 \includegraphics[height=\eyew ]{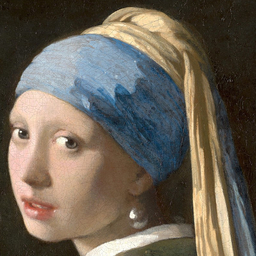} &
 \includegraphics[height=\eyew ]{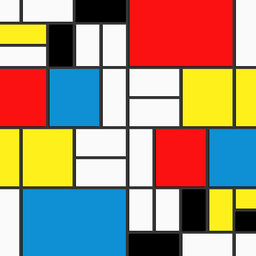} &
 \includegraphics[height=\eyew ]{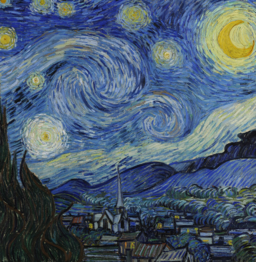} &
  \includegraphics[height=\eyew ]{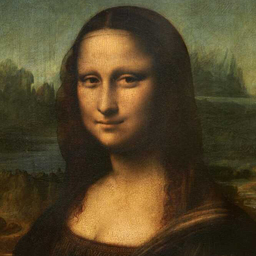}&
 \includegraphics[height=\eyew ]{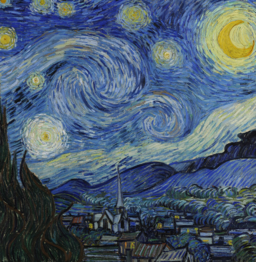} &
 \includegraphics[height=\eyew ]{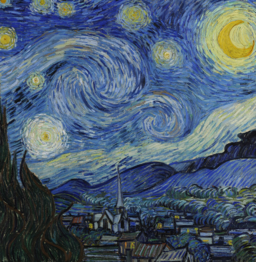} \\
 &

   \multicolumn{4}{c|}{\footnotesize $32\times32$} & 
   \multicolumn{2}{c|}{\footnotesize$16\times16$} & 
\footnotesize $8\times8$ \\

 \hline 
 & & & & & & & \\

 \includegraphics[width=0.27in]{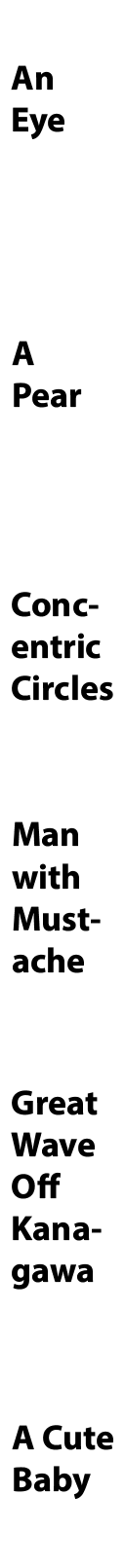}&
 \includegraphics[width=\eyew ]{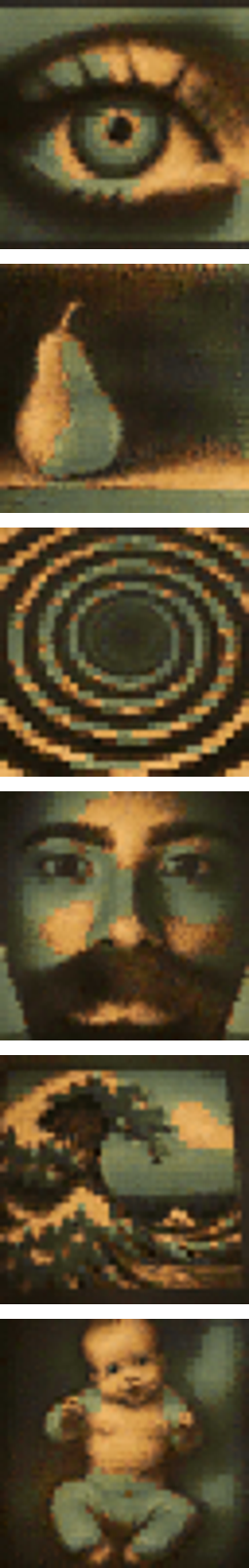} &
 \includegraphics[width=\eyew ]{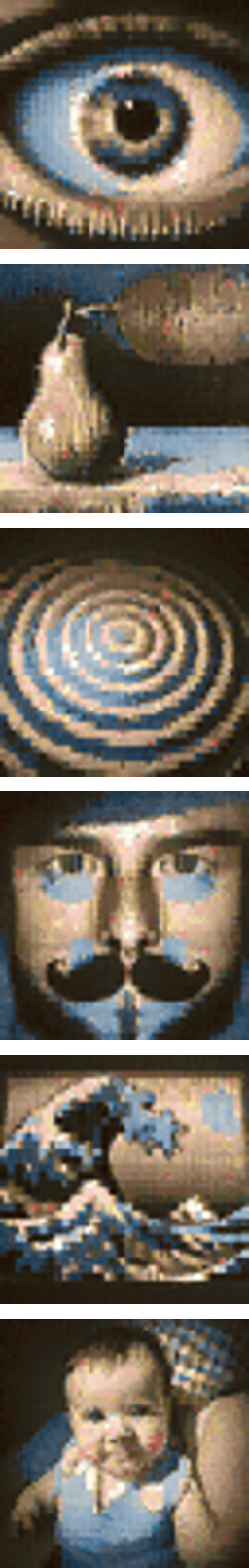} &
 \includegraphics[width=\eyew ]{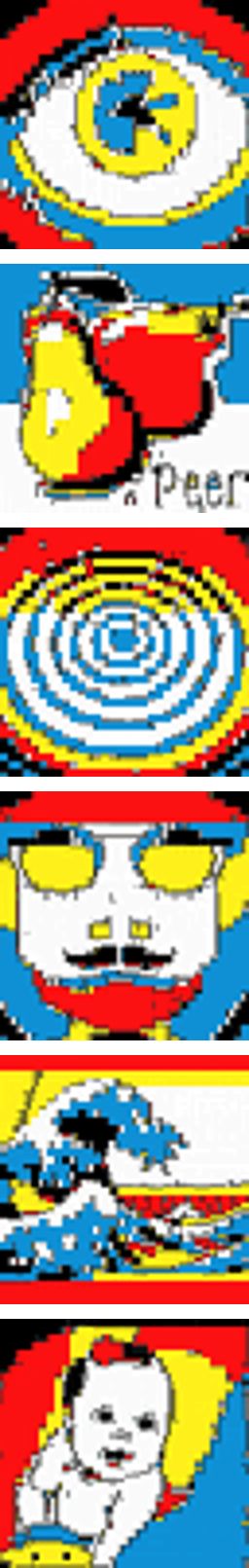} &
 \includegraphics[width=\eyew ]{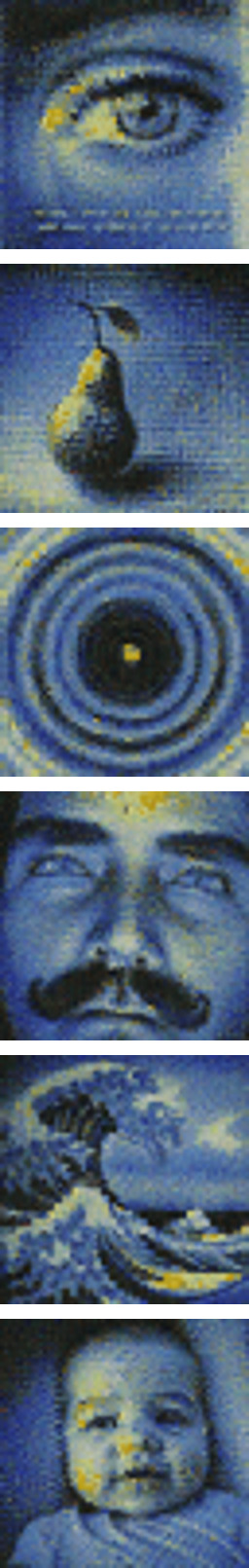} &
 \includegraphics[width=\eyew ]{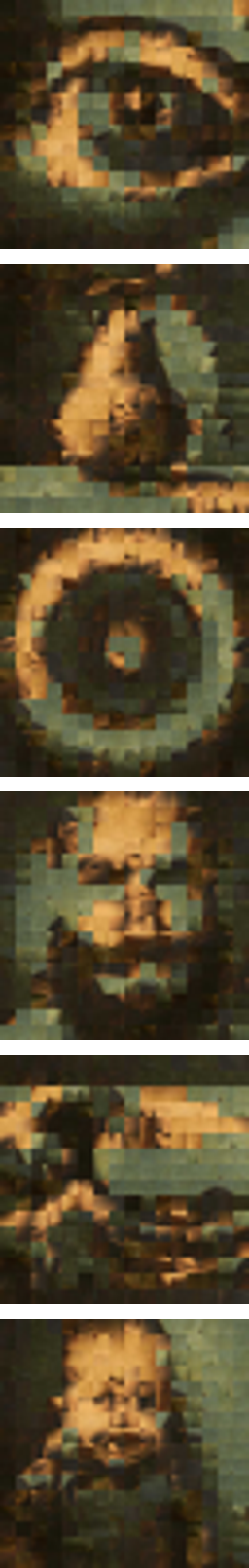}&
       \includegraphics[width=\eyew ]{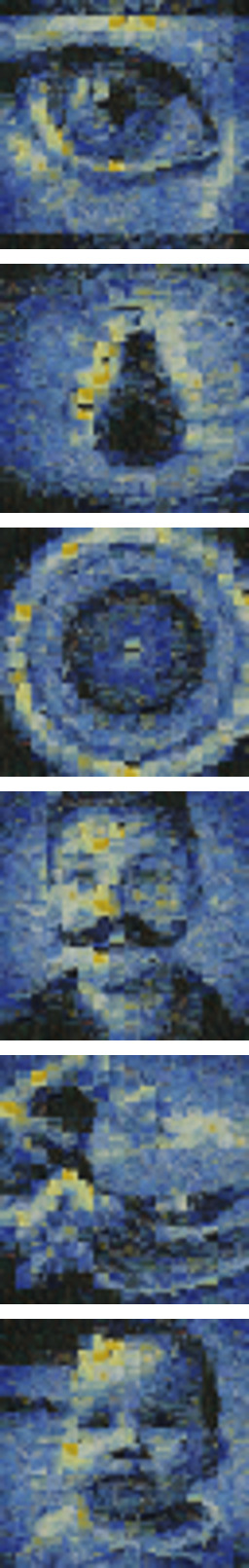}&
    \includegraphics[width=\eyew ]{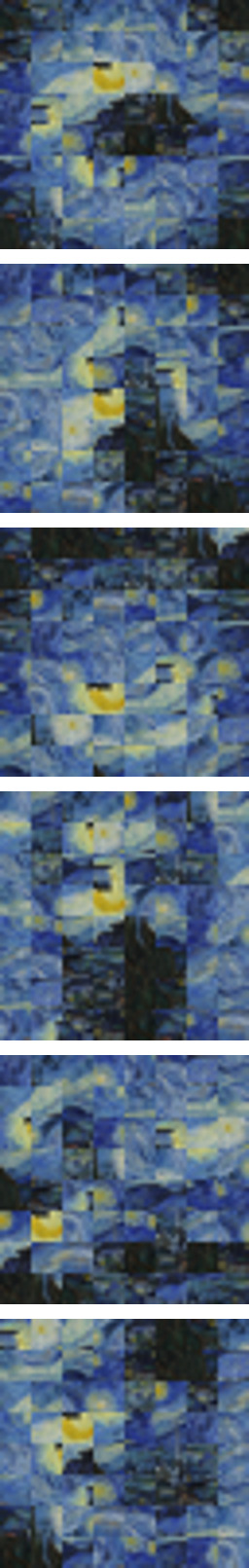}\\

 \end{tabular}
 \caption{Examples using fixed source images. Top Row is the original image ($\beta$), the bottom 6 rows shows the images created by permuting the tiles in $\beta$.  Examples are shown for $32\times32$ to $8\times8$ tiles;  the exact same pixels as the original are used.}
 \label{fig:imagesAsInput}
\end{figure}

To use a source image, $\beta$, as the basis for the image creation,  we start with the method described in the previous section.  Since we no longer need to determine the actual values of the pixels (only where the pixels will be moved from the source), we can simplify the process.  
Within the context of the procedures shown thus far, we set the first image to $\beta$ and the prompts start at $i=2$; see Figure~\ref{fig:fixedDFAlgorithm}.    The salient differences between the previous procedures and the current are: 

\begin{enumerate}
\item $\psi^1_t$ is set to the identity function. $x_t$ is no longer an input since all rollout inputs can be derived as $\Psi_t(\beta)$.

\item Given $\psi_t$, we directly compute $\Psi_t(\beta)$ for initialization. Recall that  $\Psi^i_t(\beta)$ is a permutation of a noise-free source image.
For a rollout to be useful (since rollouts use diffusion internally), it expects a noisy image, not one that is noise-free.  Therefore, we \emph{re-add} noise to $\Psi^i_t(\beta)$ to create the rollout input, a process termed ``rollout mixing''.  We can characterize rollout mixing as:
\begin{equation}
w_t^{image}\Psi^i_t(\beta) + w_t^{noise}\epsilon
\end{equation}
where $\epsilon \sim \mathcal{N}(0,1)$ is noise.  $w_t^{image}$ and $w_t^{noise}$ for step $t$ are computed directly from the diffusion schedule (see~\cite{Ho2020denoising,rombach2021highresolution} for details).
An effect of the diffusion schedule is that, the earlier in the mainline process, the more weight is given to $w_t^{noise}$, allowing larger changes.

\item We perform each rollout (see Figure~\ref{fig:fixedDFAlgorithm}) to completion, which yields an idealized noise-free image that we call $Q^i_t$.  The tiles in $Q_t$ are matched with $\beta$ revealing the new $\psi_{t-1}$.  This is analogous to the matching process earlier, where we matched to the first idealized image.

\item Once dynamic matching has computed the new assignment of tiles, there is no further need to adjust the pixel values. They are directly taken from $\beta$.  The Visual Anagram step is no longer required. 

\end{enumerate}

Results with square crops of four paintings (da Vinci's \emph{Mona Lisa},
Vermeer's \emph{Girl with a Pearl Earring}, a Mondrian-like artwork, and
Van Gogh's \emph{Starry Night}) used as the source image, $\beta$, are
shown in Figure~\ref{fig:imagesAsInput}.  Each artwork is converted to
6 new subjects.  When $\beta$ contains a variety of natural shading
artifacts, performance is vastly improved.  Conversely, when $\beta$
has few colors and no shading (column 3), the results are more akin to
clip-art.  As we increase the number of tiles, quality improves. The
use of a predetermined source image significantly amplifies the
constraints, thereby exacerbating the complexity of the problem; fewer
tiles than $16 \times 16$ produced inferior images.  One of the
subtle, but most interesting, findings in these results is that since
the same seed was used to generate these images, the differences in
the images of each row is due to the constraints imposed by matching
the source image.  A visualization of how rollouts progress is shown
in the Appendix. 

To this point, we have demonstrated our approach with two prompts.
However, that is not a limitation of the procedure.  In fact, with the use of a $\beta$
image, it can be shown that $N$ prompts need not be solved
simultaneously. Instead, equivalent results will be achieved when,
using $\beta$, each of the $N$ prompts is solved independently.  This
will yield images that are transforms of each other.

\section{Working in Latent Spaces}
\label{sec:latentDiffusion}

To this point, we have used DeepFloyd to allow direct comparison
to~\cite{geng2023visualanagrams}, but its pixel-based representation
of $x$ is uncommon in modern diffusion systems.  Since latent
diffusion-based systems are rapidly progressing in quality, we have
also extended our system to work with latent diffusion systems,
specifically Stable Diffusion's version 2.1 (SD
2.1)~\cite{rombach2022high}.  Rather than operating directly on
pixels, the diffusion of $x_t$ is in a more compact ``latent''
representation that can be encoded and decoded from/to pixels by using
a pre-trained encoder and decoder, both of which are a core part of
all pre-trained latent-diffusion systems.

Upon first glance, the simplest approach in using a latent diffusion
system would be to simply translate all the mechanisms with DeepFloyd
directly, and work entirely in latent space, with a decode only for
the final output.  However, latent diffusion introduces two significant
challenges.  First, as witnessed in~\cite{geng2023visualanagrams},
transforms on the latent representation may not preserve spatial
characteristics.  This problem is exacerbated by dynamic matching,
which also occurs in the pixel-space.  Second, latent diffusion is
hyper-sensitive to early stages of denoising; rolling out from intermediate
steps produces little change in the idealized image.

We closely follow the steps in Section~\ref{sec:fixed} and introduce modifications as necessary.  
As before, the rollout mixing step adds noise before the
start of each rollout.  Each rollout proceeds 50 steps, $S=50$. (While
50 steps is more than we used with DeepFloyd, we empirically found
this to be necessary with latent diffusion.)  We keep the mainline
steps at $T=15$, sufficient to reach convergence. As before, the
output of each rollout is labeled $Q^i_t$ and it is a noise-free
idealized image.

Experimentally, we found that latent diffusion solidifies many
characteristics of the final image in the very early diffusion
steps. Starting a rollout with higher fractions of the image (as
opposed to noise), as we did with DeepFloyd, causes output $Q_t$ to be
too similar to the rollout input.  This prevents rollouts from
creating novel images and eliminates their usefulness. To allow the
rollout enough freedom to create new images, we always start rollouts from the beginning of
the diffusion process (\emph{i.e.} $s=50 \rightarrow s=0$).  To do
this, we start with rollout mixing:
$w_S^{image}\Psi_t^i(x_t)+w_S^{noise}\epsilon$
using a ``mixing ratio'' $\frac{w_S^{image}}{w_S^{image}+w_S^{noise}}$ of $2\%$.
At $2\%$, the image provides enough direction when combined with the noise to influence the rollout, but not too much so as to limit the rollout's ability to change the input. Numerous other values were tried;  ranges of $1-4\%$ were most often successful. 

\begin{figure}[t]
  \includegraphics[width=\linewidth]{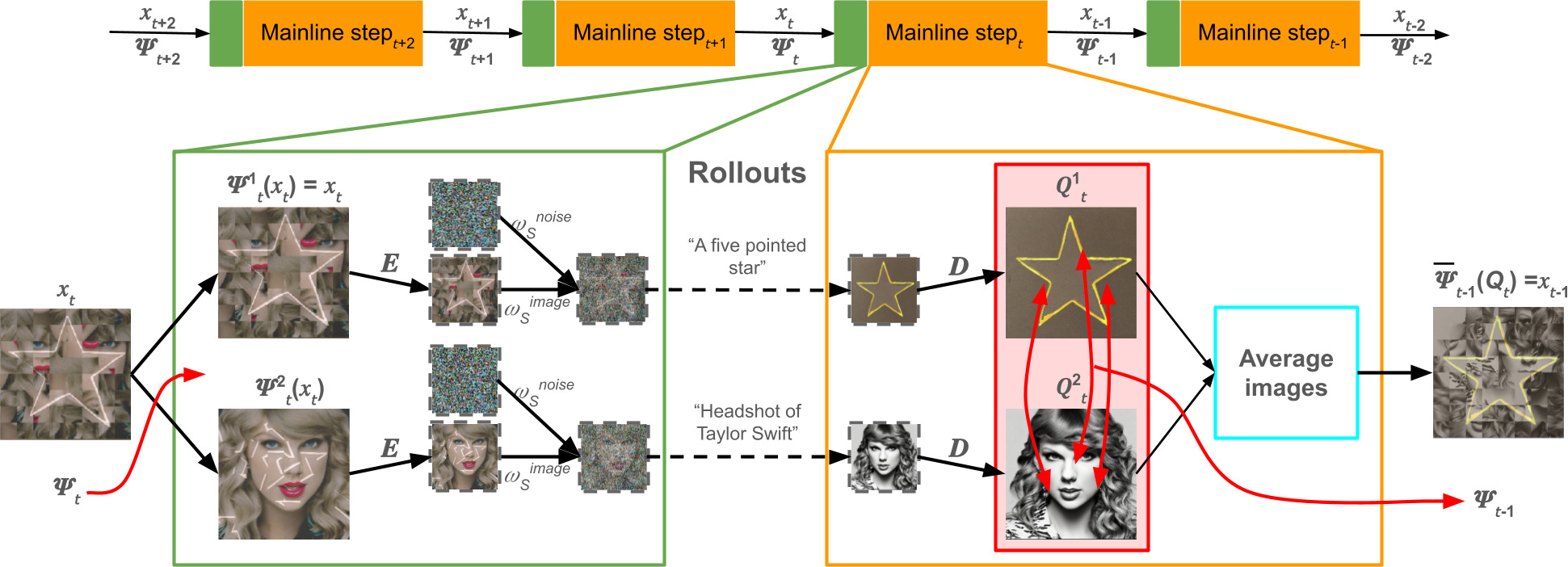}  
  
 \caption{\textbf{Changes for latent diffusion}. Latent representations, with a dashed outline, are used for diffusion and pixel images are used for dynamic matching and calculating $\Psi$. Latents are decoded with $D$ for dynamic matching then re-encoded with $E$; $D \& E$ are pretrained components of the latent diffusion system.   Rollout mixing uses fixed $w_S^{image}$ and $w_S^{noise}$. Rollouts span a full diffusion run from $s=50$ to 0.}
 \label{fig:latentProcess}
\end{figure} 

Once the idealized image has been created, it is again tempting to  simply translate
all the mechanisms with DeepFloyd directly, and work entirely in
latent space, with only a decode for the final output.  However, the
dynamic matching is spatial over pixels -- \emph{i.e.} in the decoded
image space.  To perform the matching, we must, therefore, decode the output of
each rollout to produce the pixels for $Q$, the input to dynamic
matching.  See Figure~\ref{fig:latentProcess}.  \footnote{Note, despite being ``latent'' features, their commonly employed spatial layouts lends themselves to encoding spatially localized information.  Therefore, we conducted experiments performing the transformations directly in the latent spaces.  However, these \emph{did not} produces as compelling results.}

We  note that it is important to consider when to add noise
(Figure~\ref{fig:latentProcess}). One choice is to add noise to the
pixel image and then encode; however,
$E(\Psi^i_t(w_S^{image}x_t+w_S^{noise}\epsilon))$ doesn't guarantee
the noise component is preserved as $\mathcal{N}(0,1)$ in latent
space.  Instead, we add noise \emph{after} encoding.  The input to each
rollout is therefore: $w_S^{image}E(\Psi_t^i(x_t)) + w_S^{noise}\epsilon$.

An additional implementation detail: when using dynamic matching, the
first image $Q^1_t$ is used to produce $\psi_{t-1}$ (see red box in
Figure~\ref{fig:latentProcess}).  We experimented with using
$\Psi^1_{t-1}(Q^1_t)$ as the output of the mainline step.  However, we
found that using an average of the rollout outputs worked a bit
better:
\begin{equation}
    \bar{\Psi}_{t-1}(Q_t) = \psi^1_{t-1}(\frac{1}{N}\sum_{i=1}^N \psi_t^{i^{-1}}(Q^i_t))
\end{equation}

This may be due to the specifics of the diffusion process (SD 2.1) and
may be switched back when other systems are employed.

\begin{figure}
 \newcommand{\ldhC}{0.75in}
 \newcommand{\ldh}{0.70in}
 \centering
 \tiny
 \setlength{\tabcolsep}{1pt} %
 \begin{tabular}{>{\raggedright\arraybackslash}p{\ldhC} 
                 >{\raggedleft\arraybackslash}p{\ldhC}  
                 p{0.5in}
                 >{\raggedright\arraybackslash}p{\ldhC}                  
                 >{\raggedleft\arraybackslash}p{\ldhC}} 
  \includegraphics[height=\ldh]{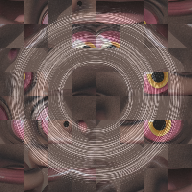} &
  \includegraphics[height=\ldh]{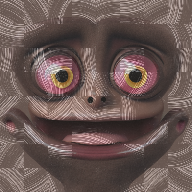} &
  ~ &
  \includegraphics[height=\ldh]{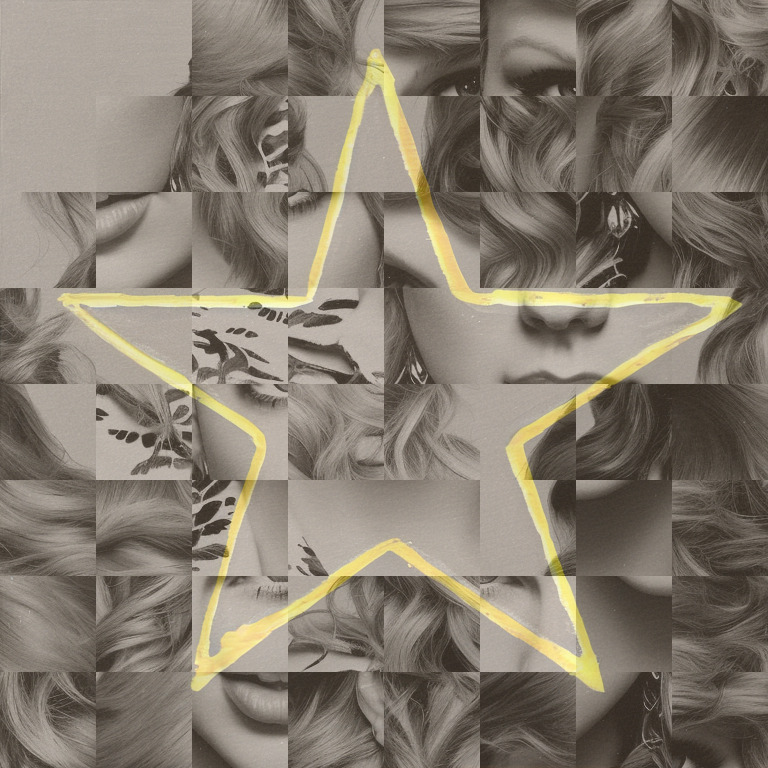} &
  \includegraphics[height=\ldh]{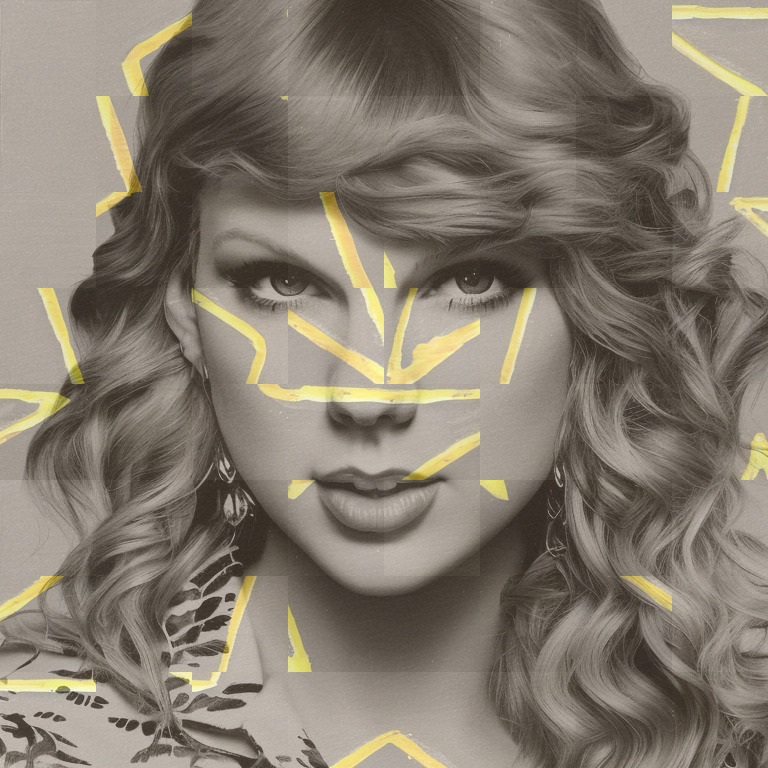} \\
  a circle on a dark background & face with big eyes and open mouth & ~ &  
  a five pointed star & headshot of Taylor Swift \\

  \includegraphics[height=\ldh]{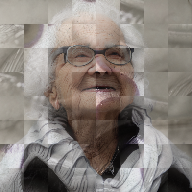} &
  \includegraphics[height=\ldh]{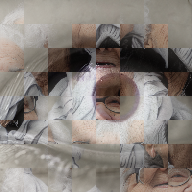} &
  ~ &
  \includegraphics[height=\ldh]{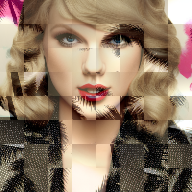} &
  \includegraphics[height=\ldh]{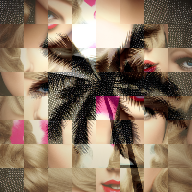} \\  
  a grandma & a closeup of an eye & ~ &
 headshot of Taylor Swift & Palm Tree\\

  \includegraphics[height=\ldh]{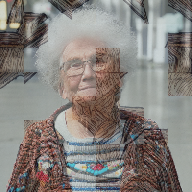} &
  \includegraphics[height=\ldh]{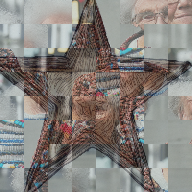} &
  ~ & 
  \includegraphics[height=\ldh]{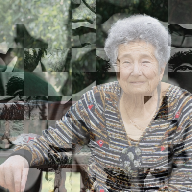} &
  \includegraphics[height=\ldh]{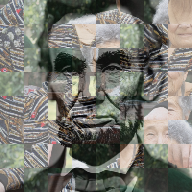} \\
  a grandma & a five pointed star & ~ &
  a grandma & headshot of Abraham Lincoln 

 \end{tabular}
 \caption{\footnotesize Examples created using latent diffusion. The output images are $768 \times 768$ pixels. The left of each image pair is divided into an $8 \times 8$ grid of tiles that are permuted to get the image on the right.  We repeat the same prompt with different pairs to show the effects of the constraints imposed by the parallel diffusion processes. }
  \label{fig:latentExamples}
\end{figure}

Examples created using latent diffusion are shown in
Figure~\ref{fig:latentExamples}. Perhaps the most evident difference
between these images and the results with DeepFloyd is in the quality
improvement.  The latent diffusion model tends to produce clearer
images than DeepFloyd, with far less structured noise and more
distinct edges.
The overall improved quality of the
images with respect to the fidelity to the prompt is reflected in the
lower CLIP distances: 21.0 for the (4x4) and 21.0 for the (8x8); both are
statistically significantly lower ($p<0.01$) than the results with
DeepFloyd.

\begin{figure}
\scriptsize
 \newcommand{\mih}{0.87in}
 \centering
 \setlength{\tabcolsep}{1pt} %
 \begin{tabular}{lp{\mih}p{\mih}p{\mih}p{\mih}p{\mih}}
  & Marilyn &          &             & George     & Abraham \\
  & Monroe  & Einstein & Shakespeare & Washington & Lincoln \\
  \rotatebox{90}{\begin{minipage}{\mih}\centering
  $16 \times 16$ tiles
  \end{minipage}} &
  \includegraphics[height=\mih]{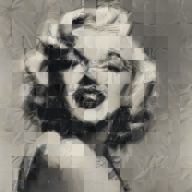} &
  \includegraphics[height=\mih]{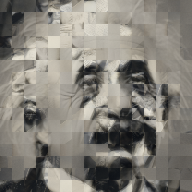} &
  \includegraphics[height=\mih]{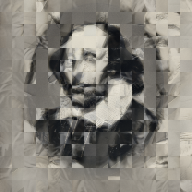} &
  \includegraphics[height=\mih]{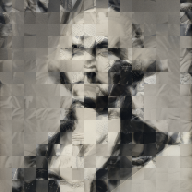} &
  \includegraphics[height=\mih]{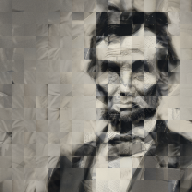} \\

  \rotatebox{90}{\begin{minipage}{\mih}\centering
  $32 \times 32$ tiles
  \end{minipage}} &
  \includegraphics[height=\mih]{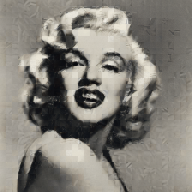} &
  \includegraphics[height=\mih]{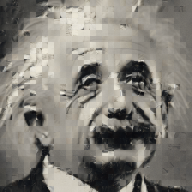} &
  \includegraphics[height=\mih]{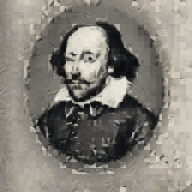} &
  \includegraphics[height=\mih]{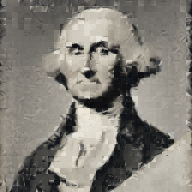} &
  \includegraphics[height=\mih]{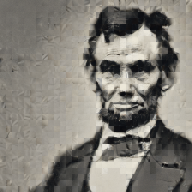} \\

  \rotatebox{90}{\begin{minipage}{\mih}\centering
  $64 \times 64$ tiles
  \end{minipage}} &
  \includegraphics[height=\mih]{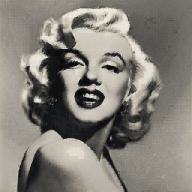} &
  \includegraphics[height=\mih]{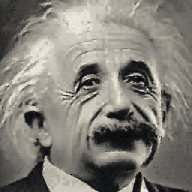} &
  \includegraphics[height=\mih]{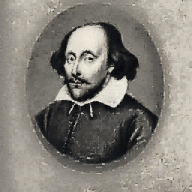} &
  \includegraphics[height=\mih]{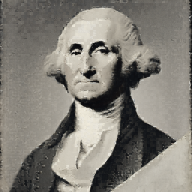} &
  \includegraphics[height=\mih]{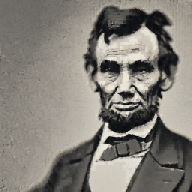} \\

  \rotatebox{90}{\begin{minipage}{\mih}\centering
  $32 \times 32$ tiles\\No matching
  \end{minipage}} &
  \includegraphics[height=\mih]{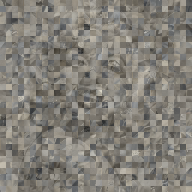} &
  \includegraphics[height=\mih]{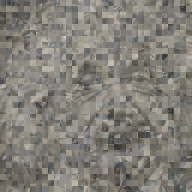} &
  \includegraphics[height=\mih]{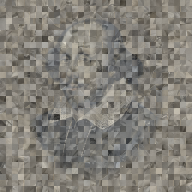} &
  \includegraphics[height=\mih]{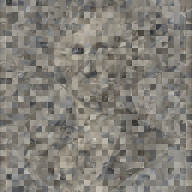} &
  \includegraphics[height=\mih]{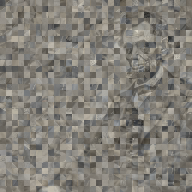} \\

  \rotatebox{90}{\begin{minipage}{\mih}\centering
  $16 \times 16$\\Vis. Anagrams
  \end{minipage}} &
  \includegraphics[height=\mih]{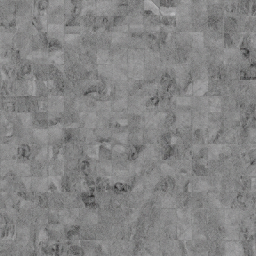} &
  \includegraphics[height=\mih]{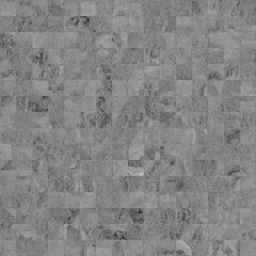} &
    \includegraphics[height=\mih]{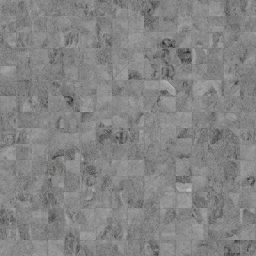} &
  \includegraphics[height=\mih]{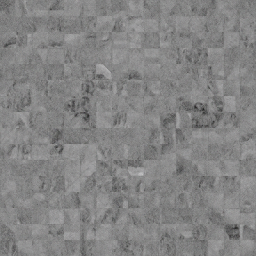} &
  \includegraphics[height=\mih]{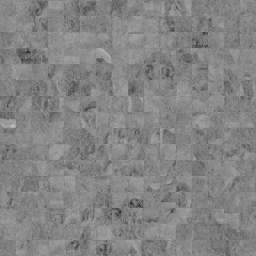} \\

  \rotatebox{90}{\begin{minipage}{\mih}\centering
  $32 \times 32$\\Vis. Anagrams
  \end{minipage}} &
  \includegraphics[height=\mih]{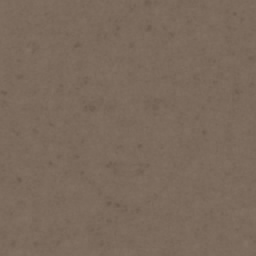} &
  \includegraphics[height=\mih]{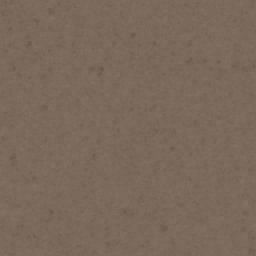} &
    \includegraphics[height=\mih]{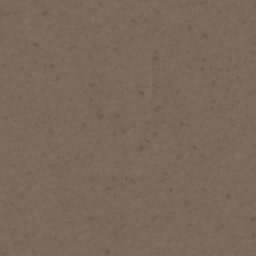} &
  \includegraphics[height=\mih]{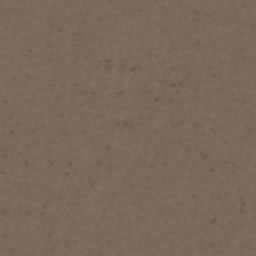} &
  \includegraphics[height=\mih]{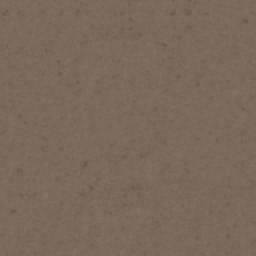} \\
 \end{tabular}
 
 \caption{ \textbf{Multiple prompts comparison}. The images in a row use the same tiles. The images are $768 \times 768$ pixels and the tiles are $48 \times 48$, $24 \times 24$ or $12 \times 12$ pixels in image rows 1, 2 and 3, respectively. ``No matching'' skips the dynamic matching. ``Vis. Anagrams'' uses the implementation from~\cite{geng2023visualanagrams}. To show each systems at their best, the Visual Anagrams system prepended ``a painting'', and ours prepended ``a photograph of''. 
Most interestingly,  without dynamic matching, there is only a barely  perceptible resemblance to the prompt; instead, the easiest matches are large fields of consistent color, with only minute variations to reflect the prompt.  
  In contrast, with dynamic matching, the images appear bright and easily recognizable.     
}
 \label{fig:multipleImages2}
\end{figure}

\begin{figure}[ht]
\scriptsize
 \newcommand{\mih}{0.85in}
 \centering
 \setlength{\tabcolsep}{1pt} %
 \begin{tabular}{lp{\mih}p{\mih}p{\mih}p{\mih}p{\mih}}
  & a house  & a sports car & a fruit & a firetruck & a lobster \\
  \rotatebox{90}{\begin{minipage}{\mih}\centering
  $64 \times 64$ tiles
  \end{minipage}} &
  \includegraphics[height=\mih]{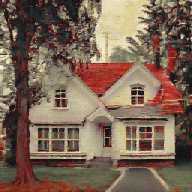} &
  \includegraphics[height=\mih]{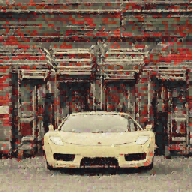} &
  \includegraphics[height=\mih]{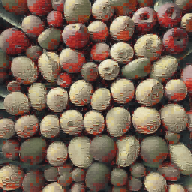} &
  \includegraphics[height=\mih]{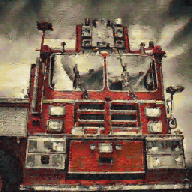} &
  \includegraphics[height=\mih]{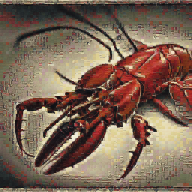} \\

  & a house  & a sports car & a fruit & a swimming pool & a diamond \\
  \rotatebox{90}{\begin{minipage}{\mih}\centering
  $64 \times 64$ tiles
  \end{minipage}} &
  \includegraphics[height=\mih]{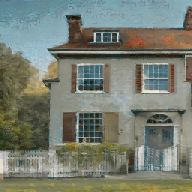} &
  \includegraphics[height=\mih]{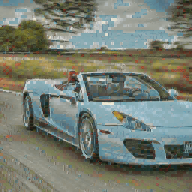} &
  \includegraphics[height=\mih]{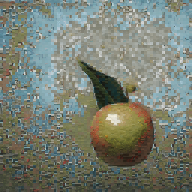} &
  \includegraphics[height=\mih]{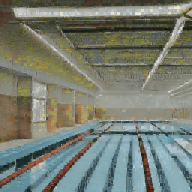} &
  \includegraphics[height=\mih]{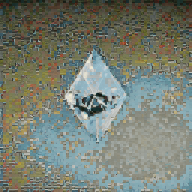} \\

  & a house  & a sports car & a fruit & a tree & an avocado \\
  \rotatebox{90}{\begin{minipage}{\mih}\centering
  $64 \times 64$ tiles
  \end{minipage}} &
  \includegraphics[height=\mih]{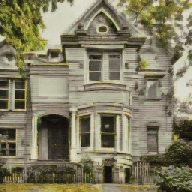} &
  \includegraphics[height=\mih]{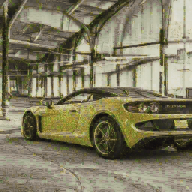} &
  \includegraphics[height=\mih]{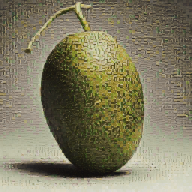} &
  \includegraphics[height=\mih]{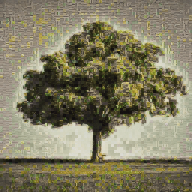} &
  \includegraphics[height=\mih]{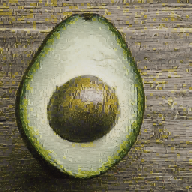} \\

 \end{tabular}
 
 \caption{\small A visual demonstrations of how constraints are propagated through the parallel diffusion processes.  In these three examples that use 5 prompts, 3 prompts are kept constant, while the last two are changed.  The last 2 prompts have strong colors associated with them.  Notice how the prototypical object's colors specified by the last 2 prompts is visible in all 5 images -- an effect of constraint propagation.}

 \label{fig:multipleImages3}
\end{figure}

Figure~\ref{fig:multipleImages2}
 gives a demonstration of how our system scales with multiple
images and large number of constraints. Unlike previous systems where
the larger number of tiles and prompts severely degrade performance,
the larger number of tiles provides our approach more degrees of
freedom for better results.  Figure~\ref{fig:multipleImages3}
also shows results with 5 prompts, with prompts chosen to present a clear visualization
of how each image affects all the others by propagating constraints (in
this case most clearly colors). 

\subsection{Known Source Image}

Analagous to Section~\ref{sec:fixed}, we describe how to use an \emph{a priori} specified source image within a latent diffusion-based system.   We continue to use the same overall settings as the latent diffusion described above:  we retain the full independent rollouts of length $S=50$ as well as the  2\% rollout mixing ratio. 

\begin{figure}
 \newcommand{\sfs}{0.7in}
 \scriptsize 
 \centering
 \def\arraystretch{2}%
 \begin{tabular}{ccc}

 \multicolumn{3}{c}{ \scriptsize $16\times16$ tiles}\\
 \midrule
 original &prompt & copies = {1,5,10} \\
 \midrule
 \midrule
  \tiny
 \multirow{2}{\sfs}[0.4in]{ \includegraphics[height=\sfs ]{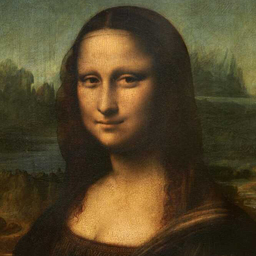}} &
 a tea cup &
 \includegraphics[height=\sfs ]{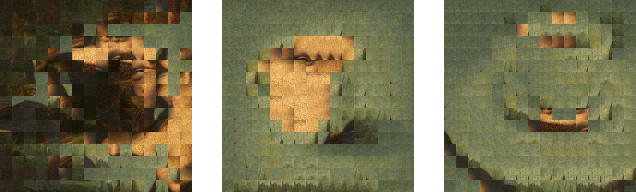} \\
 &
 
 a young boy &
 \includegraphics[height=\sfs ]{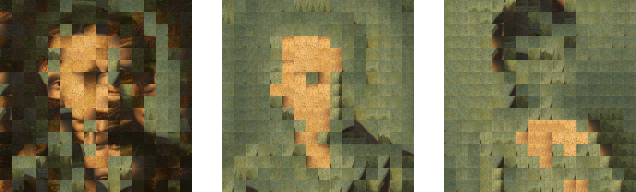}  \\
 
  \multirow{2}{\sfs }[0.4in]{ \includegraphics[height=\sfs ]{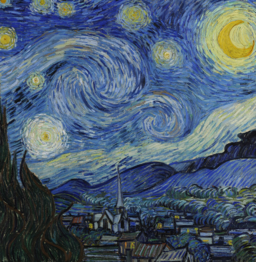}} & 
 a tea cup &
 \includegraphics[height=\sfs ]{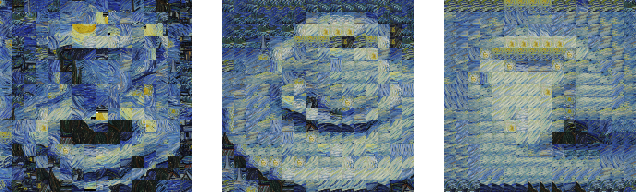} \\ 
 &
 
~~a young boy~~ &
 \includegraphics[height=\sfs ]{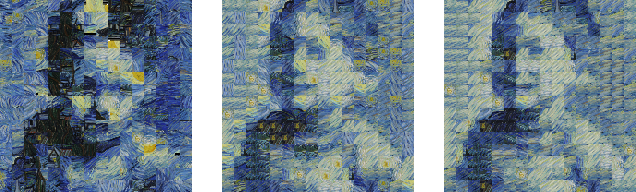} \\ 

\end{tabular}

~\\
~\\
~\\

 \begin{tabular}{ccc}
 \multicolumn{3}{c}{$32\times32$ tiles}\\
 \midrule
 \multirow{2}{\sfs }[0.4in]{ \includegraphics[height=\sfs ]{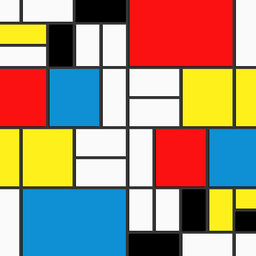}} &
 Great Wave &
 \includegraphics[height=\sfs ]{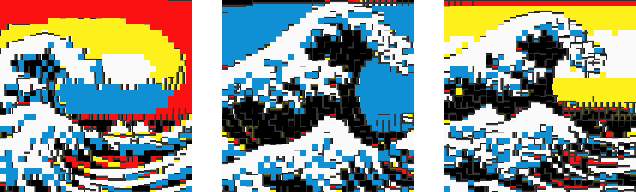} \\
 &
 
 a magpie &
 \includegraphics[height=\sfs ]{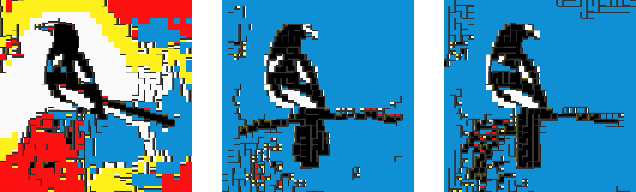}  \\

  \multirow{2}{\sfs }[0.4in]{ \includegraphics[height=\sfs ]{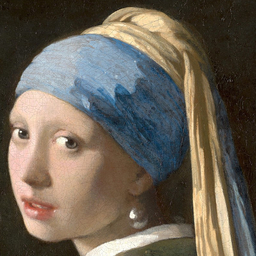}} & 
 ~~~Great Wave~~~ &
 \includegraphics[height=\sfs ]{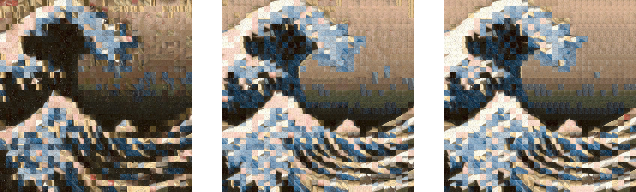} \\ 
 &

a magpie &
 \includegraphics[height=\sfs ]{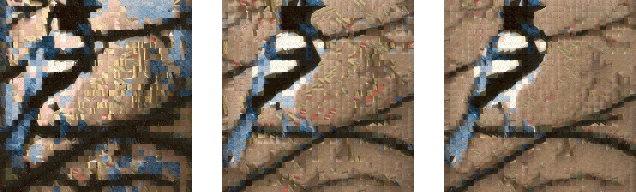} \\ 

 \end{tabular}
 \caption{Any number of image copies can be used as the basis.  In each of the examples, we look at 1,5 and 10 copies of the original to create the prompt. 
 Top Table: $16\times16$ tiles and prompts: of ``a tea cup'' and ``a young boy'' created with the \emph{Mona Lisa} and \emph{Starry Night}.  Bottom Table: $32\times32$ tiles and prompts of: ``a magpie'' and ``The Great Wave off Kanagawa'' created with Mondrian-like squares and \emph{Girl with the Pearl Earring}.}
 \label{fig:copies2}
\end{figure}

In order to accommodate the source image, we note that $Q^1$ is set to
$\beta$.  As earlier, there is no rollout required for the source
image.  Importantly, unlike Section~\ref{sec:fixed}, where we
\emph{switched} to explicitly adding noise back into $\Psi^2$
(Figure~\ref{fig:fixedDFAlgorithm}), we note that the latent diffusion
system already uses this approach, so this requires no further change.
The final outputs are the images $\Psi_0(x_0)$.  Results are shown in
Figure~\ref{fig:copies2} under copies=1.

\section{Creative Extensions}

\subsection{Infinite Images and Multiple Basis Images}
\label{sec:infiniteImages}

In Section~\ref{sec:permutations}, we characterized the tile
assignment as a perfect matching.  However, when using a fixed source
image, we can consider an interesting variant -- what if we had $c$
(where $c$ could be infinite) copies of the source image, $\beta$, and
could select the tiles to use from any of the copies?  In this
scenario, as before, every tile in the permuted image must be assigned
to a tile in $\beta$, but not every tile in $\beta$ needs to be used.
This can be characterized as a matching in a bipartite graph $(S, T;
E)$ where vertices $S$ contains $c$ vertices per fixed image tile and $T$ contains the tiles of the permuted image.  The solution is a (non-perfect) matching on the graph such that every vertex in $T$ is
part of the match.

Revisiting Equation~\ref{eq:dynamicMatching}, let $D_c$ be a $cM^2
\times M^2$ matrix in which the the rows are duplicated $c$ times. The
$M^2 \times cM^2$ assignment from the permuted image to the source is,
as before, a (0,1)-matrix $P_c$ with exactly one 1 in each row and at
most one 1 in each column. We still have:

\begin{equation}
    \min_{P_c} ~tr(P_cD_c)
\end{equation}

Here, we define the trace of a non-square matrix $A$ with shape
$(a_1,a_2)$ as $tr(A)=\sum_{i=1}^{min(a1,a2)} a_{ii}$. The solution is
implemented as the non-square variant of the Kuhn-Munkres algorithm.
Results are shown in Figure~\ref{fig:copies2} (see column copies=5 \&
copies=10). Finally, as a further extension to this approach, we note
that $c$ can be set \emph{per tile}, thereby allowing fine-grained
control over the specific tiles that should appear more often in the
results.

A straightforward variation of using multiple copies of the same image
as the source material is using multiple \emph{different} images as
the source material.  This easily fits into the framework presented in
this section with minimal changes; selections of tiles from any of the basis images are allowed for each position in the synthesized image.

\subsection{Alternate Transformations}
\label{sec:alternateTransforms}

To this point, we have focused upon the \underline{Permutation}
transformation. This transformation worked particularly well since using
the Kuhn–Munkres algorithm provided a known, and efficient, method for
matching tiles in multiple images.

Here, we demonstrate that the procedure of interleaving diffusion's
denoising step with energy minimization matching is general and works
across other transformations.  We will demonstrate this using
\underline{Concentric Rings} and \underline{Flips}.  For both, we are
able to use simple greedy and efficient matching procedures.

\begin{figure}
\centering
  a.~\includegraphics[height=1in]{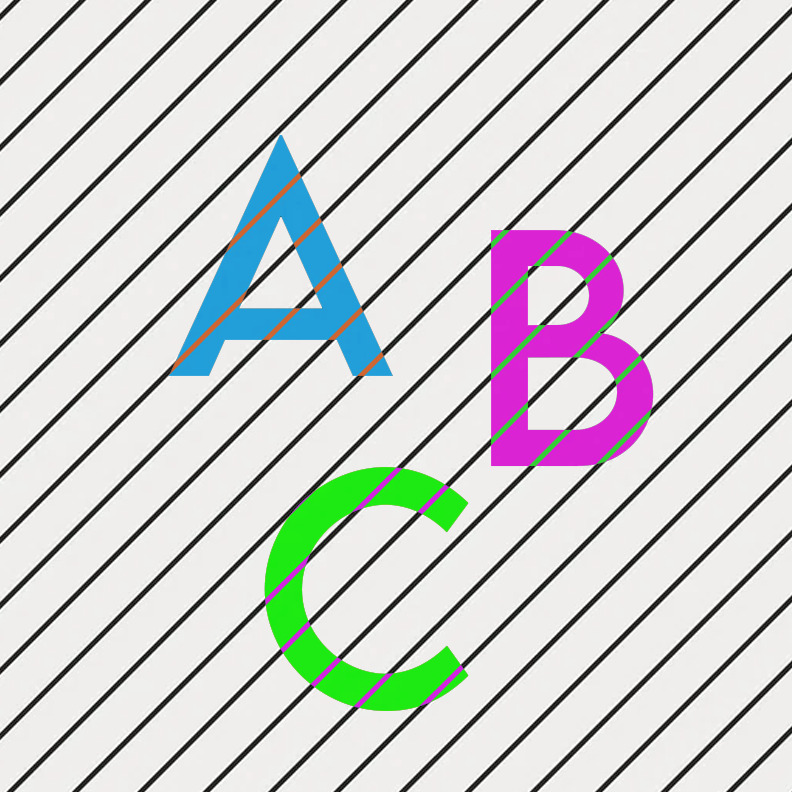} 
  ~~~~b.~\includegraphics[height=1in]{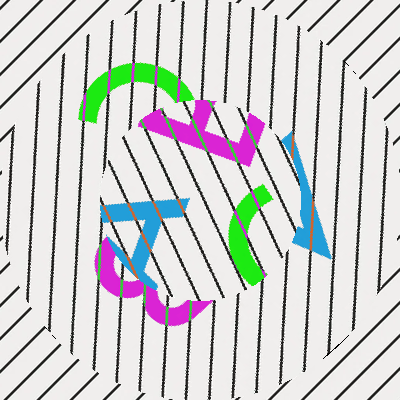}
  ~~~~c.~\includegraphics[height=1in]{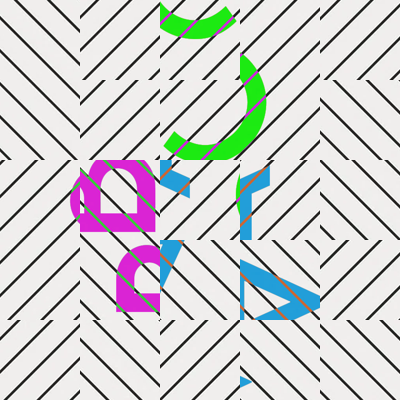} 

  \caption{Two transforms are shown here.  The first is the \emph{Concentric Rings transformation}. Here, concentric circles are rotated independently; the original, shown in (a), is rotated with 2 concentric circles, shown in (b).   Second, the \emph{Flips transformation}, allows for horizontal flips and 4-cardinal rotations of blocks at various resolutions in the image pyramid.  In (c), the image is flipped at the largest resolution and various flips and rotations are done independently for various blocks at the $5 \times 5$ division level.}
  \label{otherTransforms}
\end{figure}

In the Concentric Rings transformation, we place $C$
concentric rings, each with equal radial size, around the center of the
image.  The rings can be rotated independently to transform image A
into image B, see
Figure~\ref{otherTransforms}b.  Analogously to the work in the
previous sections, the rotations of the rings are not pre-specified.
Instead, the best rotation of the rings is computed at every mainline
step using the rolled-out images.  Differences between corresponding 
rings in image A and B are computed and their sum is minimized.

\begin{figure}
  \newcommand{\sfs}{0.68in}
  \scriptsize 
  \centering

  \begin{tabular}{lccccc}

  & Picasso's drawing of a & ~ & a graphic postcard of the & ~ & a painting of \\
  & [face, horse] & ~ & [Eiffel Tower, Taj Mahal] & ~ & [an old man, a campfire] \\
 
  \rotatebox{90}{\begin{minipage}{\sfs}\centering
  Permutation 8x8 
  \end{minipage}} &
  \includegraphics[height=\sfs ]{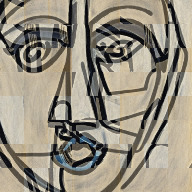} 
  \includegraphics[height=\sfs ]{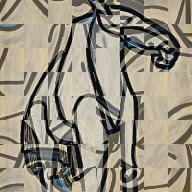} &
  ~ &
  \includegraphics[height=\sfs ]{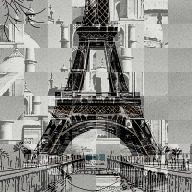} 
  \includegraphics[height=\sfs ]{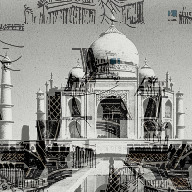} &
  ~ &
  \includegraphics[height=\sfs ]{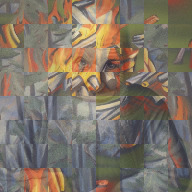} 
  \includegraphics[height=\sfs ]{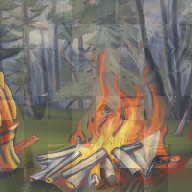} \\

  \rotatebox{90}{\begin{minipage}{\sfs}\centering
  Permutation 16x16 
  \end{minipage}} &
  \includegraphics[height=\sfs ]{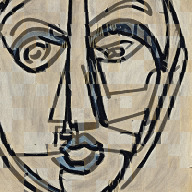} 
  \includegraphics[height=\sfs ]{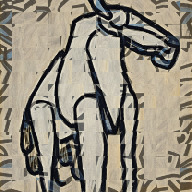} &
  ~ &
  \includegraphics[height=\sfs ]{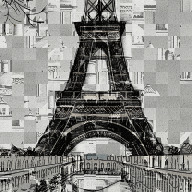} 
  \includegraphics[height=\sfs ]{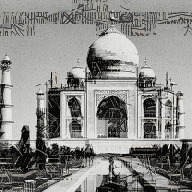} &
  ~ &
  \includegraphics[height=\sfs ]{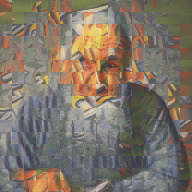} 
  \includegraphics[height=\sfs ]{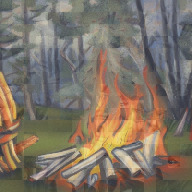} \\

  \rotatebox{90}{\begin{minipage}{\sfs}\centering
  Permutation 32x32
  \end{minipage}} &
  \includegraphics[height=\sfs ]{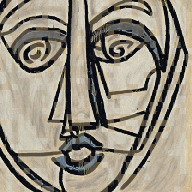} 
  \includegraphics[height=\sfs ]{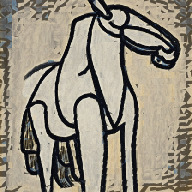} &
  ~ &
  \includegraphics[height=\sfs ]{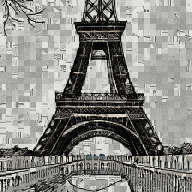} 
  \includegraphics[height=\sfs ]{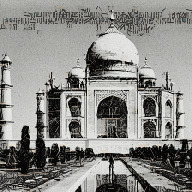} &
  ~ &
  \includegraphics[height=\sfs ]{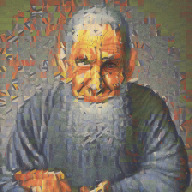} 
  \includegraphics[height=\sfs ]{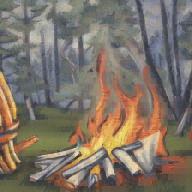} \\

  ~ \\

  \rotatebox{90}{\begin{minipage}{\sfs}\centering
  Rings \\ C=8
  \end{minipage}} &
  \includegraphics[height=\sfs ]{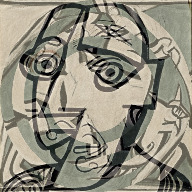} 
  \includegraphics[height=\sfs ]{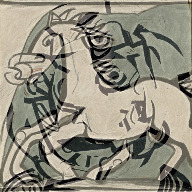} &
  ~ &
  \includegraphics[height=\sfs ]{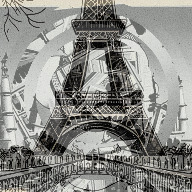} 
  \includegraphics[height=\sfs ]{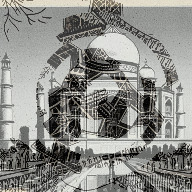} &
  ~ &
  \includegraphics[height=\sfs ]{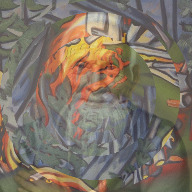} 
  \includegraphics[height=\sfs ]{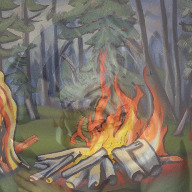} \\

  \rotatebox{90}{\begin{minipage}{\sfs}\centering
  Rings \\ C=16 
  \end{minipage}} &
  \includegraphics[height=\sfs ]{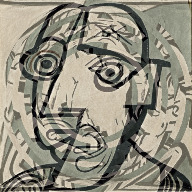} 
  \includegraphics[height=\sfs ]{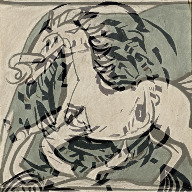} &
  ~ &
  \includegraphics[height=\sfs ]{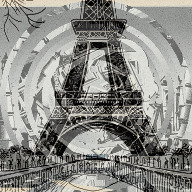} 
  \includegraphics[height=\sfs ]{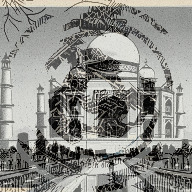} &
  ~ &
  \includegraphics[height=\sfs ]{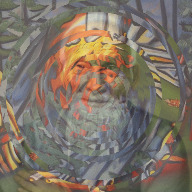} 
  \includegraphics[height=\sfs ]{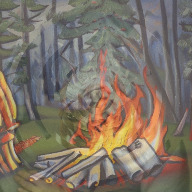} \\
  
  \rotatebox{90}{\begin{minipage}{\sfs}\centering
  Rings \\ C=25 
  \end{minipage}} &
  \includegraphics[height=\sfs ]{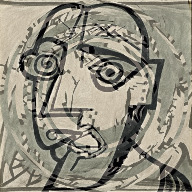} 
  \includegraphics[height=\sfs ]{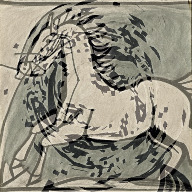} &
  ~ &
  \includegraphics[height=\sfs ]{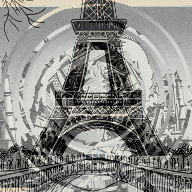} 
  \includegraphics[height=\sfs ]{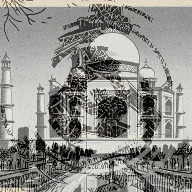} &
  ~ &
  \includegraphics[height=\sfs ]{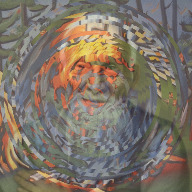} 
  \includegraphics[height=\sfs ]{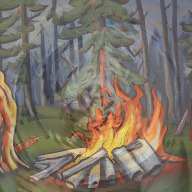} \\
  
  ~ \\

  \rotatebox{90}{\begin{minipage}{\sfs}\centering
  Flips D=(1,10)
  \end{minipage}} &
  \includegraphics[height=\sfs ]{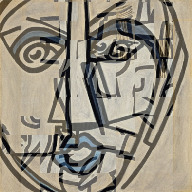} 
  \includegraphics[height=\sfs ]{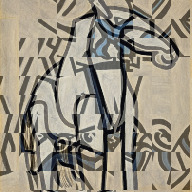} &
  ~ &
  \includegraphics[height=\sfs ]{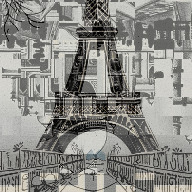} 
  \includegraphics[height=\sfs ]{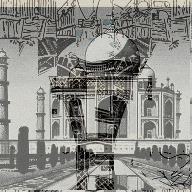} &
  ~ &
  \includegraphics[height=\sfs ]{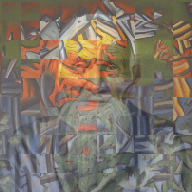} 
  \includegraphics[height=\sfs ]{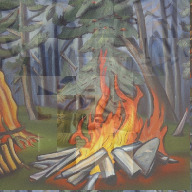} \\

  \rotatebox{90}{\begin{minipage}{\sfs}\centering
  Flips D=(1,16)
  \end{minipage}} &
  \includegraphics[height=\sfs ]{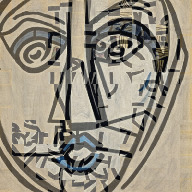} 
  \includegraphics[height=\sfs ]{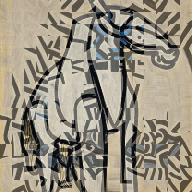} &
  ~ &
  \includegraphics[height=\sfs ]{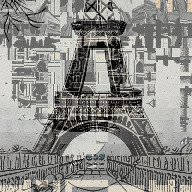} 
  \includegraphics[height=\sfs ]{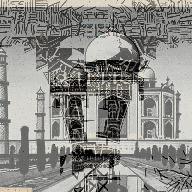} &
  ~ &
  \includegraphics[height=\sfs ]{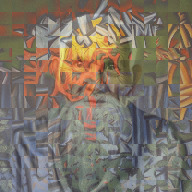} 
  \includegraphics[height=\sfs ]{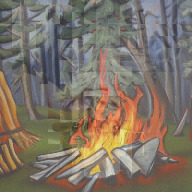} \\
  
\end{tabular}
  \caption{Results for the Permutation, Concentric Rings and Flips transformations using similar prompts. Horizontal pairs are transforms of one another using the transformation in the left column. }
  \label{otherTransformsExamples}
\end{figure}

Results are shown in Figure~\ref{otherTransformsExamples}, middle three rows.  Here, we evaluated 
72 rotations for each ring (in 5$^{\circ}$ increments) independently of the other rings.  The larger the number of rings that were used, the clearer the results.  

For the \underline{Flip} transformation, we experimented with tile
rotation and mirroring effects at multiple levels in the image
pyramid.  Given a division, $d$, we divide the original square image
into $d \times d$ equal sized blocks.  The allowed manipulations for
each block are a flip (horizontally) and rotation by
(0$^{\circ}$, 90$^{\circ}$, 180$^{\circ}$, 270$^{\circ}$).  Rather than
doing this at a single level, we allow this to happen at several
divisions, $D$.  For example, a division set $D=(1,5)$, where
the full image can be modified as well as at the $\frac{1}{5}$ boundaries, might yield a result as shown in
Figure~\ref{otherTransforms}c.

Results are shown in Figure~\ref{otherTransformsExamples}, bottom two rows.  For simplicity, we
only allowed two resolutions of operation: the entire image, and the
lowest resolution (shown in each row).  This first provides a
mechanism to make a global change, and the second a mechanism for
smaller, local, changes.

Though it is difficult to make quantitative, general, statements on these qualitative results, we provide a few observations.  In general, the original Permutation transformation provided the most compelling results.   The permutations of tiles has two attributes that we suspect are important.  First, the permutation itself provides a substantial amount of freedom of movement.  Blocks can be moved throughout the image, thereby allowing very different images to be created from the same set of pixels.  Second, the movement keeps together most spatially localized pixels.  As natural images exhibit a high amount of spatial cohesiveness, moving entire blocks preserves patches of consistent colors and textures.   

The Concentric Rings transformation does substantially better with more rings.  However, note that each ring rotation (especially in the outer rings) controls the position of spatially disparate pixels and the region of a single ring manipulation may extend the full width of the image.  This necessitates creating images using potentially unnatural constraints.   

The Flips transformation, on the other hand, maintains spatial cohesiveness.  In the examples shown, we allowed flips \& rotations at full and at either $8\times 8$ or $16\times 16$ blocks.  In our initial trials, we experimented with not including the full level transformation. If the top level was omitted, the system often failed in creating two recognizable distinct images. In the extreme, although very fine-grained tiling adds parameters to the transformation, it results in excessively local changes that make it impossible to create noticeably different images -- there is an empirical compromise between $d=1$ global changes and very fine-grained tiling. Allowing the $d=1$ global change also helped ameliorate  locality effects.   Nevertheless, the constraints on this transformation are far more restrictive than on the Permutation transformation. 

\section {Conclusions \& Discussion}

Simply by rearranging an image's tiles, we presented methods to transform an image into a novel one of any subject matter.  This method extends and improves recent work in generation of optical illusions in four fundamental aspects.

\begin{enumerate}

\item \textbf{The arrangement of tiles need not be \emph{a priori}
  specified.}  Rather, through a dynamic matching step, the
  arrangement is jointly created with the image. Beyond creating
  qualitatively and quantitatively better images, this opens new
  functionality.

\item \textbf{A source image can be pre-specified.}  Previously, all
  images had to be created by the process to allow their content to be
  matched.  Dynamic matching removes that constraint, allowing a
  user-specified source image.

\item \textbf{The problem becomes \emph{easier} as the number of tiles
  grows}; the opposite was true with static matching.

\item \textbf{An infinite number of copies} of the source image can be
  used in matching.  Multiple different images can also be used for the source.

\end{enumerate}

We have extended this study to parameterized transforms beyond
randomized-patch permutations.  The interleaving of energy
minimization and de-noising applies equally well to a variety of
transforms.  

Looking forward, we note that though this work
presented results on a vision-processing / graphics task, in a broader
sense, the system was a method to navigate a system of constraints.
Understanding the use of the diffusion process to search through the valid
spaces in this potentially alternate category of problems remains an
exciting prospect for future work.

\newpage
\section {Appendix}
\begin{figure}[h]
  \centering
  \includegraphics[width=\textwidth]{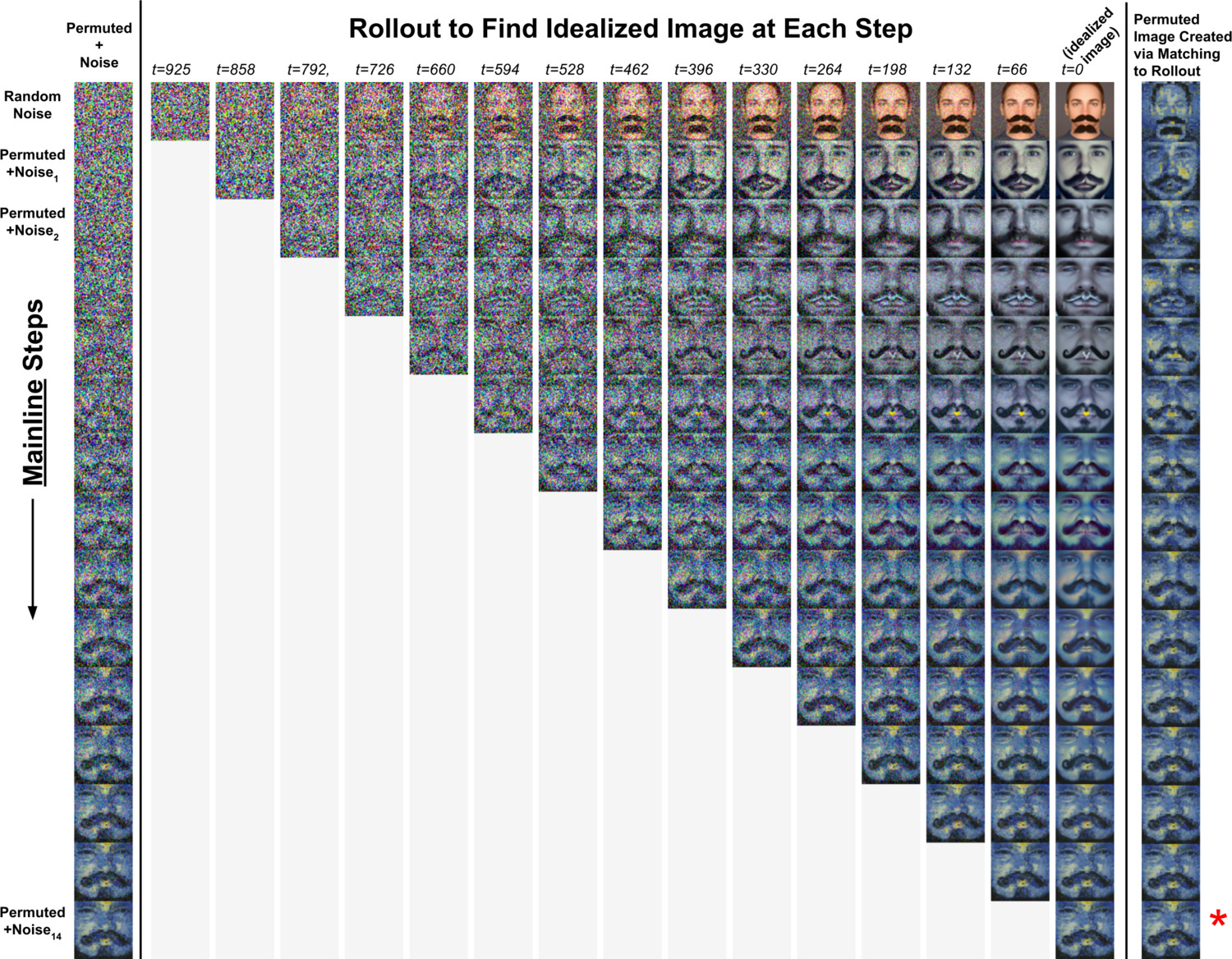}
  \caption{Rollout visualization. The mainline progress is shown on the left, progressing vertically downwards.  For each mainline step, a rollout that begins with the current mainline image computes the idealized image for the prompt, ``A Man with a Mustache''.  Once the idealized image is found, dynamic matching finds the best permutation $\psi_t$ that matches the $16\times16$ blocks from the idealized image to the $\beta$ image, \emph{Starry Night} by Van Gogh.  Noise is added as shown in Figure~\ref{fig:fixedDFAlgorithm}. The process is then repeated.  The final image is shown with a red asterisk.}

  \label{rolloutVisual}
\end{figure}

\newpage

\bibliographystyle{splncs04}
\bibliography{refs}

\end{document}